\documentclass[10pt,twocolumn,letterpaper]{article}

\usepackage[pagenumbers]{cvpr} 

\usepackage{animate}
\usepackage{graphicx}
\usepackage{amsmath}
\usepackage{amssymb}
\usepackage{booktabs}
\usepackage[accsupp]{axessibility}
\usepackage{multirow}
\usepackage{color, colortbl}
\definecolor{Gray}{gray}{0.9}
\usepackage[pagebackref,breaklinks,colorlinks]{hyperref}

\usepackage[capitalize]{cleveref}

\crefname{section}{Sec.}{Secs.}
\Crefname{section}{Section}{Sections}
\Crefname{table}{Table}{Tables}
\crefname{table}{Tab.}{Tabs.}

\def\cvprPaperID{252} 
\def\confName{CVPR}
\def\confYear{2023}

\begin{document}
	
	\title{
	
		Global-to-Local Modeling for Video-based 3D Human Pose and Shape Estimation
	}
	
	\author{
		Xiaolong Shen$^{1,2*}$, Zongxin Yang$^{1}$, Xiaohan Wang$^1$, Jianxin Ma$^2$, Chang Zhou$^2$, Yi Yang$^1$ \\
		$^1$ ReLER, CCAI, Zhejiang University \space \space \space $^2$ DAMO Academy, Alibaba Group \\
	}
	
	\maketitle

	\begin{abstract} 
		Video-based 3D human pose and shape estimations are evaluated by intra-frame accuracy and inter-frame smoothness.
		Although these two metrics are responsible for different ranges of temporal consistency, existing state-of-the-art methods treat them as a unified problem and use monotonous modeling structures (\textit{e.g.}, RNN or attention-based block) to design their networks. However, using a single kind of modeling structure is difficult to balance the learning of short-term and long-term temporal correlations, and may bias the network to one of them, leading to undesirable predictions like global location shift, temporal inconsistency, and insufficient local details. 
		To solve these problems, we propose to structurally decouple the modeling of long-term and short-term correlations in an end-to-end framework, Global-to-Local Transformer (GLoT). 
		First, a global transformer is introduced with a Masked Pose and Shape Estimation strategy for long-term modeling. The strategy stimulates the global transformer to learn more inter-frame correlations by randomly masking the features of several frames.
		Second, a local transformer is responsible for exploiting local details on the human mesh and interacting with the global transformer by leveraging cross-attention. 
		Moreover, a Hierarchical Spatial Correlation Regressor is further introduced to refine intra-frame estimations by decoupled global-local representation and implicit kinematic constraints. 
		Our GLoT surpasses previous state-of-the-art methods with the lowest model parameters on popular benchmarks, \textit{i.e.}, 3DPW, MPI-INF-3DHP, and Human3.6M. Codes are available at \url{https://github.com/sxl142/GLoT}.
		
	\end{abstract}
	\renewcommand{\thefootnote}{*}
	\footnotetext{This work was done during an internship at Alibaba.}
	\section{Introduction}
	\label{sec:intro}
	\begin{figure}[!t]
		\centering
		\begin{subfigure}{1\linewidth}
			\centering
			\includegraphics[width=0.9\textwidth]{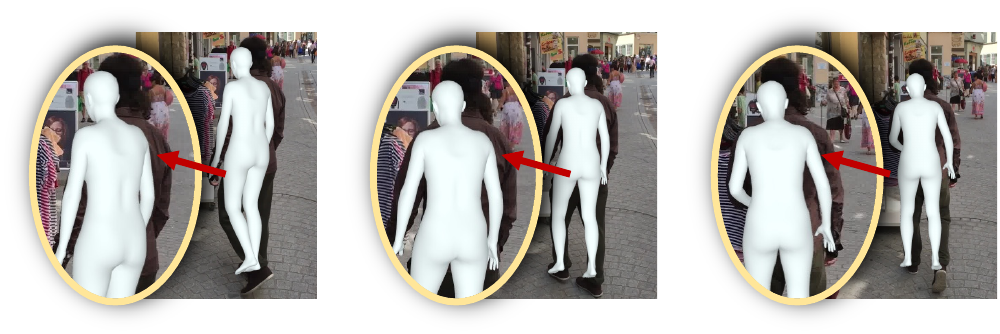}
			\vspace{-0.7 em}
			\caption{TCMR~\cite{TCMR} results. Global location shift, shifting to the left.}
			
			\label{fig:tcmr-vis}
		\end{subfigure}
		\vfill
		\begin{subfigure}{1\linewidth}
			\includegraphics[width=0.9\textwidth]{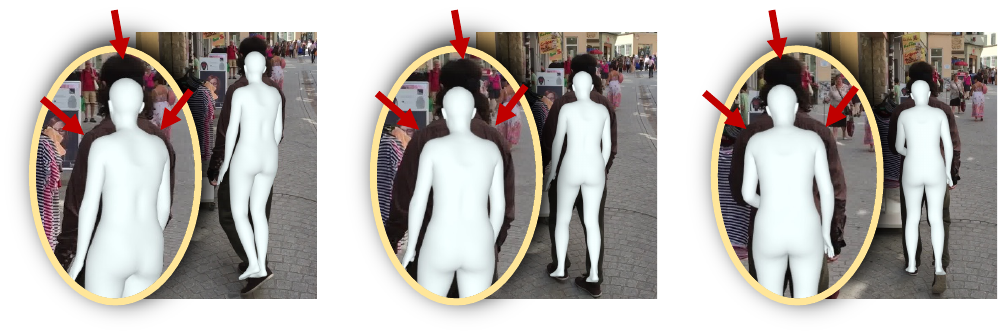}
			\vspace{-0.7 em}
			\centering
			\caption{MPS-Net~\cite{MPS-net} results. Insufficient local details.}
			\label{fig:mps-vis}
		\end{subfigure}
		\vfill
		\begin{subfigure}{1\linewidth}
			\includegraphics[width=0.9\textwidth]{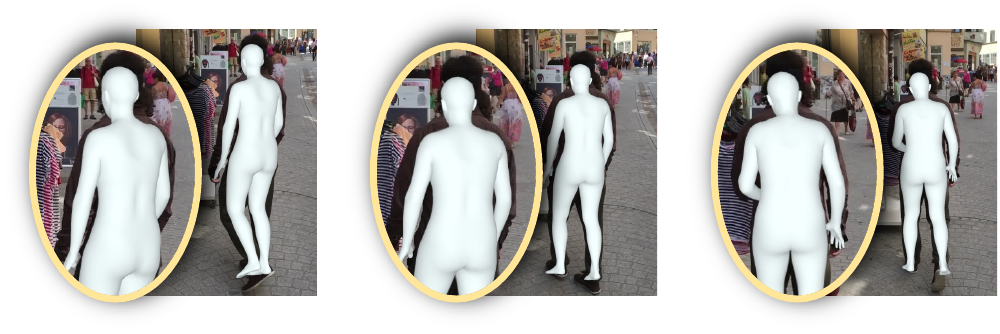}
			\vspace{-0.7 em}
			\centering
			\caption{Our results.}
			\label{fig:our-vis}
		\end{subfigure}
		\vspace{-1.7 em}
		\caption{Our motivation. With the help of global-local cooperative modeling, our results avoid the global location shift and complement local details on intra-frame human meshes.}
		\vspace{-1.8 em}
		\label{fig:motivation}
	\end{figure}


        Automatically recovering a sequence of human meshes from a monocular video plays a pivotal role in various applications, \eg, AR/VR, robotics, and computer graphics. This technology can potentially reduce the need for motion capture devices and manual 3D annotations, providing human motion templates for downstream tasks, \eg, the animation of 3D avatars. By utilizing parametric human models (\ie, SMPL~\cite{SMPL}) with well-defined artificial joint and shape structures, the popular procedure for video-based human mesh recovery involves indirectly regressing the SMPL parameters. 
        However, effectively integrating deep neural networks with parametric artificial models to leverage multi-knowledge representations~\cite{yang2021multiple} for better estimation accuracy still remains an open problem.

 In video-based human mesh recovery, temporal understanding poses a crucial challenge that necessitates maintaining both intra-frame accuracy and inter-frame smoothness. 
 Previous methods~\cite{VIBE,TCMR,MPS-net} mainly design deep networks to model long-term and short-term correlations simultaneously. For instance, VIBE~\cite{VIBE} utilizes Recurrent Neural Network (RNN)~\cite{GRU} to model correlations. TCMR~\cite{TCMR} and MPS-Net~\cite{MPS-net} consist of a temporal encoder and a temporal integration. The temporal encoder includes two types, RNN-based or attention-based methods, while the temporal integration employs attentive methods, \ie, one-step or multi-step integration, to aggregate the representation extracted by the temporal encoder.
	
	\begin{figure}[!t]
		\centering
		\includegraphics[width=0.4\textwidth]{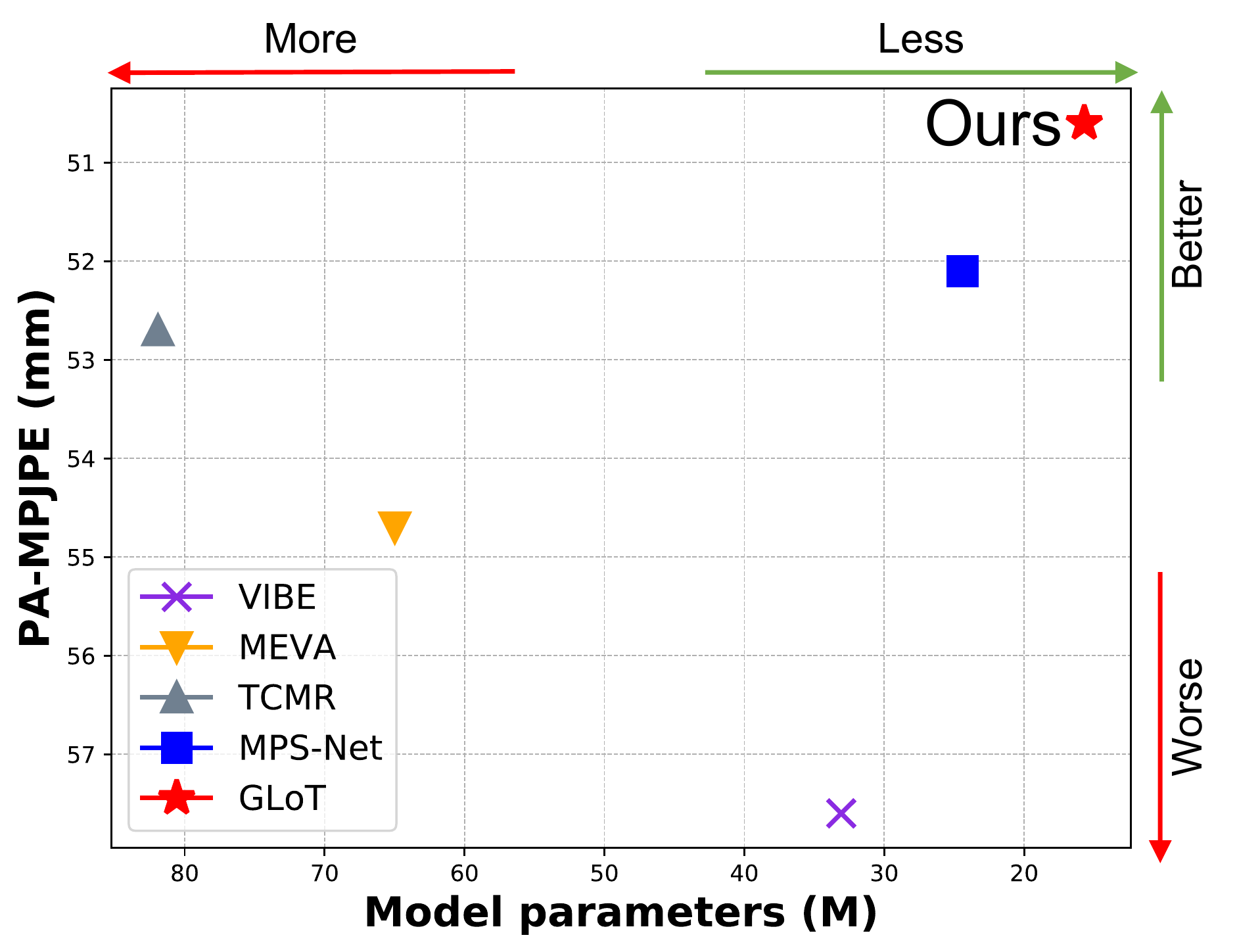}
		\vspace{-0.7 em}
		\caption{
			Model parameters vs. Performance.
		}
		\vspace{-0.8 em}
		\label{fig:para}
		\vspace{-1.0 em}
	\end{figure}
	
	Even though these methods improve intra-frame accuracy or inter-frame smoothness to some extent, designing a coupled model to handle different ranges of temporal consistency, \ie, long-term and short-term~\cite{yang2021collaborative,yang2021associating}, is ineffective and inefficient for this task, as shown in Figure~\ref{fig:para}.
	We empirically find that the single type of modeling structures leads to undesirable predictions as shown in Figure~\ref{fig:motivation}. For instance, RNN-based TCMR~\cite{TCMR} suffers from a global location shift. Attention-based MPS-Net~\cite{MPS-net} captures the proper location of human meshes in the video sequence, but the mesh shape does not fit with the human in the images.
	We consider that the coupled modeling of long-term and short-term dependencies may not balance these two sides. It leads the RNN-based method to fall into the local dependency and does not capture the global location of the human mesh in the video. In contrast, the attention-based method tends to capture the long-term dependency and does not capture sufficient local details.
	Moreover, regressing the SMPL parameters from a coupled representation also confuses the model, \eg, fine-grained intra-frame human mesh structure requires more short-term local details. 
	
        To solve the above problems, we propose to use information from both deep networks~\cite{Transformer} and human prior structures~\cite{SMPL, ZiniuWan2021EncoderDecoderWM} in a joint manner. The proposed method, namely global-to-Local Transformer (GLoT), decouples the short-term and long-term temporal modeling.
        The framework is composed of Global Motion Modeling and Local Parameter Correction. 
	In global modeling, we introduce a global transformer with a Masked Pose and Shape Estimation~(MPSE) strategy for capturing the long-range global dependency~(\eg, proper global location, motion continuity). Specifically, the strategy randomly masks the features of several frames, and then the model predicts the SMPL parameters for the masked frames, which further helps the global transformer mine the coherent consistency of human motion and guides it to seize the inter-frame correlation from a global view. 
	Under the local view, we introduce a local transformer and a Hierarchical Spatial Correlation Regressor (HSCR) for exploiting the short-term inter-frame detail and learning the intra-frame human mesh structure. 
    To achieve this, we introduce nearby frames of the mid-frame and process them through the local encoder, which utilizes the mid-frame as a query to match the global transformer encoder’s memory, generating a disentangled global-local representation of the mid-frame.
	Finally, HSCR employs human kinematic structures to constrain the refinement of decoupled global-local representation and improve global estimation.
	
	With the help of global-local cooperative modeling, our model obtains the best intra-frame accuracy and inter-frame smoothness. For example, compared with the previous state-of-the-art method~\cite{MPS-net}, our model significantly reduces the PA-MPJPE, MPJPE, and MPVPE by 1.5 $mm$, 3.6 $mm$, and 3.4 $mm$, respectively, on the widely used dataset 3DPW\cite{3DPW}, while preserving the lowest Accel metric representing the inter-frame smoothness. Moreover, our model remarkably decreases the model parameters, as shown in Figure~\ref{fig:para}.
	Our contributions can be summarized as follows:
	\begin{itemize}
		\item 
            To our best knowledge, we make the first attempt to decouple the modeling of long-term and short-term correlations in video-based 3D human pose and shape estimation. The proposed Global-to-Local Transformer (GLoT) merges the knowledge from deep networks and human prior structures, improving our method's accuracy and efficiency.
		\item 
		In GLoT, we carefully design two components, \ie, Global Motion Modeling and Local Parameter Correction, for learning inter-frame global-local contexts and intra-frame human mesh structure, respectively.
		\item 
		We conduct extensive experiments on three widely-used datasets. Our results show that GLoT outperforms the previous state-of-the-art method~\cite{MPS-net}, while achieving the lowest model parameters.
		
	\end{itemize}
	\begin{figure*}[!t]
		\centering
		\includegraphics[width=0.9\textwidth]{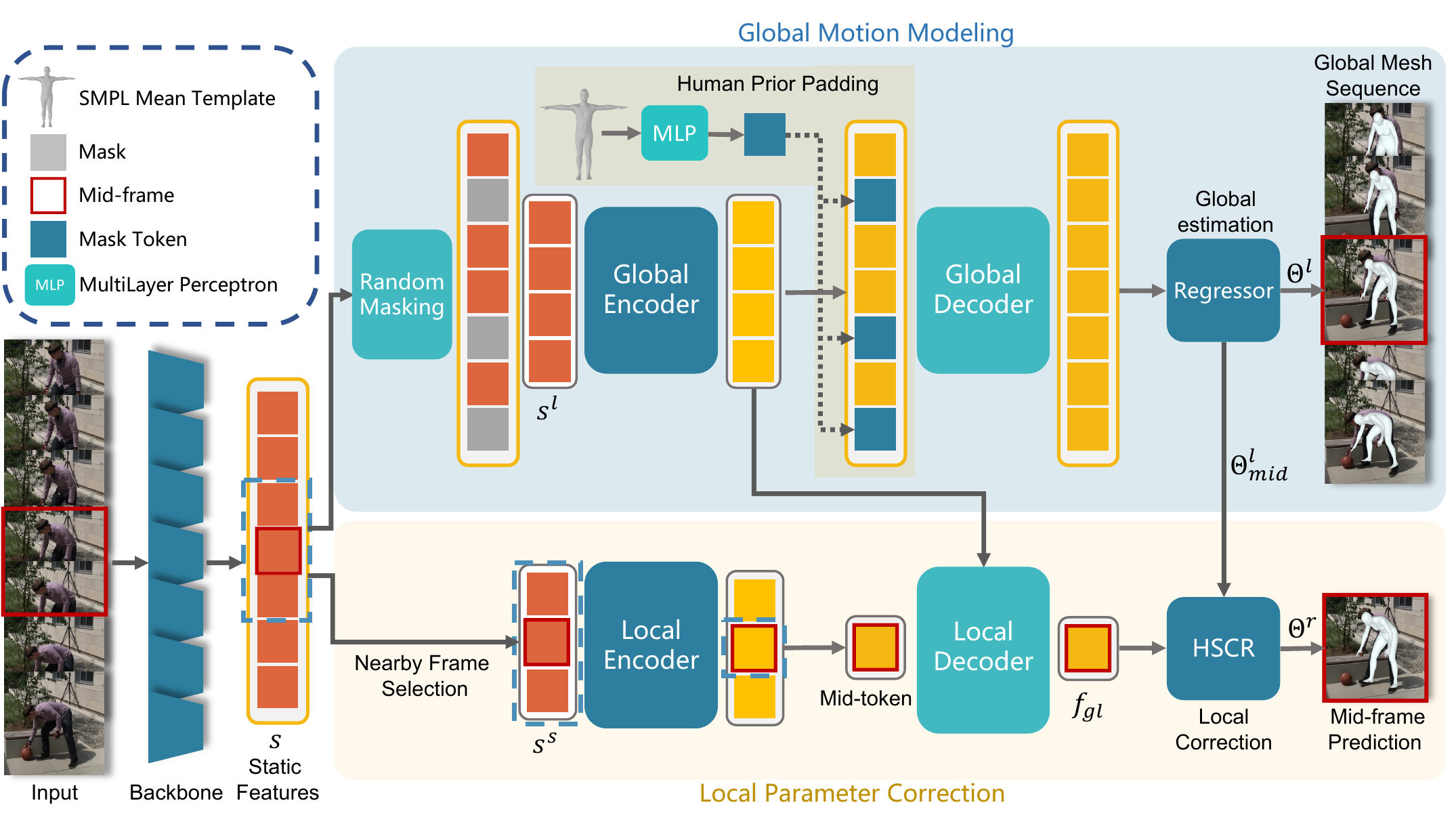}
		\vspace{-1.5 em}
		\caption{An overview of our GLoT. 
			GLoT includes two branches, \ie,  Global Motion Modeling~(GMM) and Local Parameter Correction~(LPC). We first extract static features from pretrained ResNet-50~\cite{ResNet}, following~\cite{VIBE, TCMR, MPS-net}. Then the static features $S$ are separately processed by random masking and nearby frame selection for feeding them ($S^l, S^s $) into global and local transformers. Last, Hierarchical Spatial Correlation Regressor~(HSCR) corrects the global results~$\Theta_{mid}^l$ obtained by GMM with the decoupled global-local representation $f_{gl}$ and the inside kinematic structure. Note that our method utilizes T frames to predict the mid-frame, following ~\cite{TCMR, MPS-net}.
		}
		\label{fig:Overview}
		\vspace{-1.2 em}
	\end{figure*}
	
	\section{Related work}
	\subsection{Image-based human pose and shape estimation}
	There are two lines of methods in the image-based human pose and shape estimation. The first one is the SMPL-based regression method. 
	Kanazawa \etal \cite{HMR} design an end-to-end framework named Human Mesh Recovery (HMR) for predicting the shape and 3D joint angles of the SMPL model while introducing adversarial training.
	Some methods~\cite{GeorgiosPavlakos2018LearningTE, GeorgiosGeorgakis2020HierarchicalKH, MohamedOmran2018NeuralBF, PARE, zhang2021pymaf} integrate prior knowledge, \ie, 2D joint heatmaps and silhouette, semantic body part segmentation, multi-scale contexts, and kinematic prior, with a CNN model for estimating SMPL parameters. Moreover, 3D pose estimation~\cite{GeorgiosPavlakos2016CoarsetoFineVP, MPII3D} is highly related to this task. For example, Li \etal \cite{li2021hybrik} incorporate a 3D pose estimation branch to use inverse kinematics.
	Kolotouros \etal~\cite{SPIN} unify two paradigms, \ie, optimization-based and regression-based methods, into a general framework.
	The other line~\cite{GraphCMR, moon2020i2l, metro_lin} is to directly regress human mesh without a parametric model (SMPL).
	Even though these image-based methods achieve good results, they suffer from unstable human motion when applied to video sequences. 
	
	\subsection{Video-based human pose and shape estimation}

	Video-based human pose and shape estimation mainly consists of SMPL-based regression methods. Kanazawa \etal~\cite{Insta} propose HMMR that learns a dynamics representation of humans from videos and allows the model to predict future frames. 
	Kocabas \etal~\cite{VIBE} introduce a motion prior provided by a large-scale Mocap dataset~\cite{AMASS} to guide the model in learning a kinematically reasonable human motion structure via adversarial training. Although the motion prior helps the model to capture the human kinematic structure, it still suffers from temporally inconsistent. Hence, a series of methods emerged to solve this problem.
	MEVA~\cite{MEVA} utilizes VAE~\cite{VAE} to encode the motion sequence and generate coarse human mesh sequences. The corresponding human meshes are then further refined via a residual connection. 
	TCMR~\cite{TCMR} and MPS-Net~\cite{MPS-net} can be unified into a general framework, \ie temporal encoder, and temporal integration. TCMR utilizes GRU~\cite{GRU} to encode video representation of three different input lengths and then integrates three frames, \ie, mid-frame, front and rear frames of mid-frame, with an attentive module.
	MPS-Net replaces the GRU with a Non-local based~\cite{Non_local} motion continuity attention (MoCA) module, which helps to learn non-local context relations. Moreover, instead of the one-step attentive aggregation of three frames in TCMR, MPS-Net utilizes a hierarchical attentive feature integration (HAFI) module for multi-step adjacent frames integration.
	Although these methods improve the per-frame accuracy or temporal consistency in some way, a coupled temporal modeling structure leads to some problems, \eg, global location shift, motion inconsistency, and intra-frame inaccuracy.
	
	\subsection{Vision Transformers}
	The Transformer~\cite{Transformer} is first introduced in natural language processing (NLP). Transformer-based structures are effective for capturing long-range dependency and are suitable for learning global context due to the natural property of the self-attention mechanism. 
	Inspired by the success of Transformer in NLP, Alexey \etal \cite{VIT} first propose ViT and successfully obtain good results in image classification compared to convolutional architectures. 
	Subsequently, many works~\cite{MatthijsDouze2020TrainingDI, T2T, ZeLiu2021SwinTH, zhao2022centerclip} have emerged to improve computing efficiency and enhance representation ability. 
    Moreover, recent studies~\cite{yang2021associating,yang2022decoupling,liang2023local,li2022locality,zhou2022survey} have notably employed transformers to accomplish dense video tasks~\cite{miao2021vspw,miao2022large}. 
	Except for the design of the structure, some strategies for pretraining transformers, \eg, masked Language modeling~(MLM)~\cite{BERT} in NLP, masked image modeling~(MIM)~\cite{MAE, BEIT} in CV, enhance the representation of transformers.
	
	\section{Method}
	\subsection{Overview}
	Figure~\ref{fig:Overview} shows an overview of our Global-to-Local Transformer~(GLoT), which includes two branches, namely Global Motion Modeling~(GMM) and Local Parameter Correction~(LPC).
	Given an RGB video $\textbf{I}$ with T frames, $\textbf{I} = \{I_t\}_{t=1}^{T}$, we first utilize a pretrained ResNet-50~\cite{ResNet, SPIN} to extract static features $\textbf{S} =  \{s_t\}_{t=1}^{T}, s_t \in \mathbb{R}^{2048}$. Note the static features are saved on disk as we do not train the ResNet-50. 
	 We then feed these static features into GMM and LPC to obtain the final human mesh. Next, we elaborate on each branch of this framework as follows. 

	\subsection{Global Motion Modeling}
	\label{sec:gmm}
	   The Global Motion Modeling involves three components, \ie a global transformer, a Masked Pose and Shape Estimation strategy, and an iterative regressor proposed by HMR~\cite{HMR}. For convenience in describing the overall process, we refer to the static features as static tokens. 
	
	\noindent\textbf{Global Transformer.}
	Recently, transformers have shown a powerful ability to model long-range global dependency owing to the self-attention mechanism. It is suitable for this task to capture long-term dependency and learn temporal consistency in human motion. 
	
	\noindent\textbf{Masked Pose and Shape Estimation.}
	Moreover, inspired by MIM~\cite{MAE, BEIT}, we propose a simple yet effective strategy named Masked Pose and Shape Estimation, which helps the global transformer further mine the inter-frame correlation on human motion. Specifically, we randomly mask several static tokens and predict only the SMPL parameters of the masked locations by leveraging the learned correlation with other unmasked tokens during training.
	
	\noindent\textbf{Human Prior Padding.}
	To reduce computation costs, the global transformer does not encode the masked tokens. Therefore, we need to pad tokens onto the masked location during the decoding phase. 
	We propose to use the SMPL mean template as padding for the masked tokens.
	The SMPL template represents the mean of the SMPL parameter distribution and thus has the smallest average difference to a random pose sampled from the distribution. Considering Transformers are built on a stack of residual connections, the learning will be easier from an input initialization which has a smaller residual difference from the output prediction. To handle the SMPL mean template, we utilize an MLP for dimension transformation and refer to it as an SMPL token.

	\noindent\textbf{Iterative Regressor.}
	The iterative regressor, first proposed by HMR~\cite{HMR}, has been shown to provide good estimations in some recent works~\cite{VIBE, TCMR, MPS-net}. It iteratively regresses the residual parameters of the previous step, starting from the mean SMPL parameters.
	Although the regressor may overlook some local details, \eg, human kinematic structure, it is well-suited to provide an initial global estimation under the global-to-local framework.
	
	The complete process of GMM can be described as follows, we first randomly mask some static tokens along the temporal dimension, $\textbf{S}^{l} \in \mathbb{R}^{(1-\alpha)T  \times 2048}$, $\alpha$ is a mask ratio. Then we apply the global encoder to these unmasked tokens. During the global decoder phase, following~\cite{MAE}, we pad the SMPL tokens onto the masked location and send the whole sequence into the global decoder to obtain a long-term representation. Finally, we apply the iterative regressor to the long-term representation to achieve global initial mesh sequences. We formulate the SMPL parameters obtained by GMM as follows,  $\Theta^l =\{\theta^l, \beta^l, \phi^l\}, \theta^l  \in \mathbb{R}^{T  \times (24 \times 6)}, \beta^l  \in \mathbb{R}^{T  \times 10}, \phi^l  \in \mathbb{R}^{T  \times 3}$. $\theta$ and $\beta$ are the pose and shape parameters that control the joint rotation and mesh shapes by the parametric model SMPL~\cite{SMPL}. 
	Due to insufficient 3d data annotation, $\phi$ represents a set of pseudo camera parameters predicted by the model, which project the 3d coordinates onto the 2d space for weakly supervising the model by abundant 2d annotated data.
	
	
	
	\subsection{Local Parameter Correction}
	
	Local Parameter Correction consists of two components, \ie a local transformer and a Hierarchical Spatial Correlation Regressor. 
	
	\noindent\textbf{Local Transformer.}
	As it is well known, the motion of a human body in a mid-frame is significantly influenced by its nearby frames. To capture the short-term local details in these frames more effectively, we introduce a local transformer.
	Specifically, we select the nearby frames for short-term modeling, $\textbf{S}^s = \{s_t\}_{t=\frac{T}{2} - w}^{\frac{T}{2} + w}, s_t \in \mathbb{R}^{2048}$, $w$ is the length of nearby frames. We utilize the local encoder on these selected tokens. 
	The local decoder is different from the global transformer, which decodes the features representing not only global-wise human motion consistency but also local-wise fine-grained human mesh structure through a cross-attention mechanism. The cross-attention function can be defined as follow,
	\begin{equation}
		\begin{aligned}
			CrossAtten(Q_{s}^{mid}, K_{l}, V_{l})
			&= Softmax (\frac{Q_{s}^{mid}K_{l}^T}{\sqrt[]{C}})V_{l},
			\label{eq: ca}
		\end{aligned}
	\end{equation}
	where $Q_{s}^{mid}$ is a query of the mid-token, $K_{l}$ and $V_{l}$ are key and value of the global encoder.
	
	\begin{figure}[!t]
		\centering
		\includegraphics[width=0.4\textwidth]{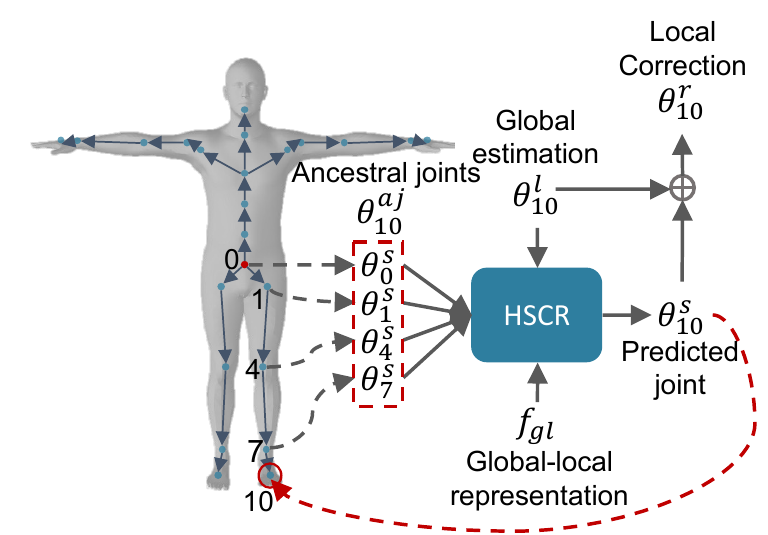}
		\caption{
			Human kinematic structure and an example of regressing and correcting one joint location.
		}
		\label{fig:kinematic}
		\vspace{-1em}
	\end{figure}
	
	\noindent\textbf{Hierarchical Spatial Correlation Regressor.}
	\label{sef:hscr}
	Previous methods~\cite{VIBE, TCMR, MPS-net} utilize the regressor from the pioneering work HMR~\cite{HMR} to estimate the SMPL parameters. However, this approach overlooks the inherent human joint correlation, \ie, kinematic structure. Although the iterative regressor can produce a good estimation of SMPL parameters, it still requires local details to create a kinematically reasonable and rendered accurate human mesh.
	
	Hence, inspired by~\cite{ZiniuWan2021EncoderDecoderWM}, we propose a Hierarchical Spatial Correlation Regressor. Considering that directly regressing $\theta$ according to the kinematic structure may cause the model to fall into sub-optimal solutions due to a local view. Therefore, we add initial global prediction and decoupled global-local representation to this regressor, which allows the model to focus on adjusting the initial global estimation from a global-to-local perspective. 
	Specifically, modeling the local intra-frame human mesh structure need to learn the joint correlation inside the kinematic structure. As shown in Figure~\ref{fig:kinematic}, the kinematic structure can be described as the current joint rotation matrix being constrained by its ancestral joints. For example, when estimating the child joint (10), we need to compute the parent joints (0, 1, 4, 7) step-by-step.
	The total process of regressing parameters can be formulated as follows,
	\begin{equation}
		\begin{aligned}
			\theta_i^{aj} &= Concat(\{\theta_{i,j}^s\}_{j=1} ^ {len(aj)}) \\
			\theta^s_i &= M_{\theta_i^s}(f_{gl}, \theta^{aj}_i, \theta^l_i ) \\
			\theta^r &= \theta^l + \theta^s, \\
			\beta^r &= \beta^l + M_{\beta}(f_{gl}, \beta^l), \\
			\phi^r &= \phi^l + M_{\phi}(f_{gl}, 	\phi^l),
			\label{eq: hscr}
		\end{aligned}
	\end{equation}
	where $M_{\theta^s_i}, M_{\beta}$ and $ M_{\phi}$ are Multilayer Perceptron (MLP), $len(aj)$ is the number of the ancestral joints of the current joint $i$, $f_{gl}$ is a decoupled global-local representation. Note $\theta$ is composed of 6D rotation representations of 24 joints.
	$\theta = \{\theta_i\}_{i=1}^{24}, \theta_i \in \mathbb{R}^6$.

	\subsection{Loss function}
	In GMM, we apply $\mathcal{L}_2$ loss to the SMPL parameters $\Theta^l$ and 3D/2D joints location, following the previous method \cite{VIBE, MPS-net, TCMR}. Note that we only compute the loss of masked location. Moreover, we empirically discover that constraints on the velocity of the predicted 3D/2D joint location can help the model learn motion consistency and capture the long-range dependency better when applying the Masked Pose and Shape Estimation strategy to the global transformer. The velocity loss can be defined as follows,
	\begin{equation}
		\begin{aligned}
			\mathcal{L}_{vel\_2d} &= \sum_{t=1}^{T-1} m_t || (jt_{2d}^{t+1} -  jt_{2d}^{t}) -  (gt_{2d}^{t+1} -  gt_{2d}^{t}) ||_2  \\
			\mathcal{L}_{vel\_3d} &= \sum_{t=1}^{T-1} m_t || (jt_{3d}^{t+1} -  jt_{3d}^{t}) - (gt_{3d}^{t+1} -  gt_{3d}^{t})||_2
			\label{eq: loss}
		\end{aligned}
	\end{equation}
	where $jt$ is a predicted 2d/3d joint location, $gt$ is the ground truth of the 2d/3d joint location. $\mathbf{M} = \{m_i\}_{i=1}^{T-1}, \mathbf{M} \in \mathbb{R}^{T-1}$ is a mask vector. $m_i$ is 1 when masking the $i$ location, otherwise, $m_i$ is 0.
	In LPC, we apply the same loss as the GMM, except that we only constrain the mid-frame.
	
	\begin{table*}[htb]
		\small
		\setlength{\tabcolsep}{1.4 pt}
		\centering
		
			\begin{tabular}{l|cccc|ccc|ccc|c}
				\toprule[2pt]
				\multirow{2}{*}{Method}& \multicolumn{4}{c|}{3DPW} & \multicolumn{3}{c|}{MPI-INF-3DHP} & \multicolumn{3}{c|}{Human3.6M} & \multicolumn{1}{c}{number of} \\
				& PA-MPJPE $\downarrow$ & MPJPE $\downarrow$ & MPVPE $\downarrow$ & Accel $\downarrow$ & PA-MPJPE $\downarrow$ & MPJPE $\downarrow$ & Accel $\downarrow$ & PA-MPJPE $\downarrow$ & MPJPE $\downarrow$ & Accel $\downarrow$ & input frames \\
				\midrule[1pt]
				\midrule[1pt]
				VIBE~\cite{VIBE} & 57.6 & 91.9 & - & 25.4 & 68.9 & 103.9 & 27.3 & 53.3 & 78.0 & 27.3 & \textbf{16} \\
				\cellcolor{Gray}MEVA~\cite{MEVA} & \cellcolor{Gray}54.7 &\cellcolor{Gray}86.9 &\cellcolor{Gray}- & \cellcolor{Gray}11.6 & \cellcolor{Gray}65.4 & \cellcolor{Gray}96.4 & \cellcolor{Gray}11.1 & \cellcolor{Gray}53.2 & \cellcolor{Gray}76.0 & \cellcolor{Gray}15.3 & \cellcolor{Gray}90 \\
				TCMR~\cite{TCMR} & 
				52.7 & 
				86.5 & 102.9& 
				7.1 & 
				63.5& 
				97.3 & 
				8.5 & 
				52.0 & {73.6} & {3.9} & 
				\textbf{16} \\
				\cellcolor{Gray}MPS-Net~\cite{MPS-net} & 
				\cellcolor{Gray}52.1 & \cellcolor{Gray}84.3 & \cellcolor{Gray}99.7& \cellcolor{Gray}7.4 & 
				\cellcolor{Gray}62.8 &\cellcolor{Gray} 96.7 & \cellcolor{Gray}9.6 &
				\cellcolor{Gray}47.4 &\cellcolor{Gray} 69.4 & \cellcolor{Gray}\textbf{3.6} & \cellcolor{Gray}\textbf{16} \\
				\textbf{GLoT(Ours)} &
				\textbf{50.6} & \textbf{80.7} & \textbf{96.3}& \textbf{6.6} & 
				\textbf{61.5} &
				\textbf{93.9} & \textbf{7.9} &
				\textbf{46.3} & \textbf{67.0} & \textbf{3.6} & \textbf{16}
				\\
				\bottomrule[1pt]
				
			\end{tabular}%
		\vspace{-0.5 em}
		\caption{
			Evaluation of state-of-the-art video-based methods on 3DPW~\cite{3DPW}, MPI-INF-3DHP~\cite{MPS-net}, and Human3.6M~\cite{H36M}. 
			All methods use 3DPW training set in training phase, but do not utilize Human3.6M SMPL parameters from Mosh~\cite{Mosh}.
			The number of input frames follows the protocol of each paper.
		}
		\vspace{-1.em}
		\label{tab:sota_video}
	\end{table*}
	\section{Implementation details}
	Following previous methods~\cite{VIBE, TCMR, MPS-net}, we set the input length $T$ to 16. and use the same data processing procedure as TCMR~\cite{TCMR}. We set the mini-batch $N$ to 64 and initialize the learning rate to 1e-4. Moreover, we apply the Cosine Annealing~\cite{IlyaLoshchilov2016SGDRSG} scheduler and linear warm-up to the Adam~\cite{DiederikPKingma2014AdamAM} optimizer. Our model is trained on one Nvidia Tesla V100 GPU. Since our model predicts the mid-frame of the input sequence, the predicted frame may fall outside the clip boundary. To handle this issue, we use nearest-padding, which duplicates the nearest boundary frame to pad the missing part. For example, when estimating frame 0, we duplicate frame 0 for 7 times. 
	More details are provided in the supplementary material.
	
	\section{Experiments}
	We begin by introducing the evaluation metrics and datasets. Next, we report the evaluation results of our model and compare them with previous state-of-the-art methods. Finally, we provide ablation studies and qualitative results to further support our findings.
	
	\noindent\textbf{Evaluation Metrics.} Following previous methods~\cite{Insta, VIBE, TCMR, MPS-net}, we report several intra-frame metrics, including Mean Per Joint Position Error~(MPJPE), Procrustes-aligned MPJPE~(PA-MPJPE), and Mean Per Vertex Position Error~(MPVPE). Additionally, we provide a result for acceleration error (Accel) to verify inter-frame smoothness.
	
	\noindent\textbf{Datasets.} The training datasets include 3DPW~\cite{3DPW}, Human3.6M~\cite{HMR}, MPI-INF-3DHP~\cite{MPII3D}, InstaVariety~\cite{Insta}, Penn Action~\cite{Penn}, and PoseTrack~\cite{PoseTrack}, following~\cite{TCMR}. We use 3DPW, Human3.6M, and MPI-INF-3DHP for evaluation. More details are provided in the supplementary material.
	
	\subsection{Comparison with state-of-the-art methods}
	\noindent\textbf{Video-based methods.}
	As shown in Table \ref{tab:sota_video}, our GLoT surpasses existing methods on 3DPW, MPI-INF-3DHP, and Human3.6M, for both intra-frame metrics~(PA-MPJPE, MPJPE, MPVPE) and inter-frame metric~(Accel). This indicates that our proposed Global-to-Local Modeling method is effective for modeling the long-range dependency~(\eg, the proper global location, coherent motion continuity) and learning the local details~(\eg, intra-frame human mesh structure).
	For example, GLoT reduces PA-MPJPE by 1.5 $mm$ , MPJPE by 3.6 $mm$, MPVPE by 3.4 $mm$, and Accel by 0.8 $mm/s^2$ compared with MPS-Net~\cite{MPS-net} on 3DPW.
	\begin{table}[t]
		\small
		\setlength{\tabcolsep}{1.2pt}
		\centering
		
			\begin{tabular}{ll|cccc}
				\toprule[2pt]
				
				\multicolumn{2}{c}{\multirow{2}{*}{Method}} & \multicolumn{4}{|c}{3DPW} \\
				& & PA-MPJPE $\downarrow$ & MPJPE $\downarrow$ & MPVPE $\downarrow$ & Accel $\downarrow$ \\
				\midrule[1pt]
				\midrule[1pt]
				\multirow{5}{*}{\rotatebox[origin=c]{90}{single image}}  & HMR~\cite{HMR} & 76.7 & 130.0 & - & 37.4 \\
				& \cellcolor{Gray}GraphCMR~\cite{GraphCMR} & \cellcolor{Gray}70.2 & \cellcolor{Gray}- & \cellcolor{Gray}- & \cellcolor{Gray}- \\
				& SPIN~\cite{SPIN} &  59.2 & 96.9 & 116.4 & 29.8 \\
				& \cellcolor{Gray}I2L-MeshNet~\cite{moon2020i2l} & \cellcolor{Gray}57.7 & \cellcolor{Gray}93.2 & \cellcolor{Gray}110.1 & \cellcolor{Gray}30.9 \\
				&PyMAF~\cite{zhang2021pymaf} & 58.9 & 92.8 & 110.1 & - \\
				\midrule[1pt]
				\multirow{5}{*}{\rotatebox[origin=c]{90}{video}}
				& HMMR~\cite{Insta} & 72.6 & 116.5 & 139.3 & 15.2 \\
				& \cellcolor{Gray}VIBE~\cite{VIBE} & \cellcolor{Gray}56.5 & \cellcolor{Gray}93.5 & \cellcolor{Gray}113.4 & \cellcolor{Gray}27.1 \\
				& TCMR~\cite{TCMR} & 55.8 & 95.0 & 111.5 & {7.0} \\ 
				& \cellcolor{Gray}MPS-Net~\cite{MPS-net} & \cellcolor{Gray}{54.0} & \cellcolor{Gray}91.6 & \cellcolor{Gray}109.6 & \cellcolor{Gray}{7.5} \\
				& \textbf{GLoT(Ours)} & \textbf{53.5} & \textbf{89.9} & \textbf{107.8} & \textbf{6.7} \\
				
				\bottomrule[1pt]
				
			\end{tabular}%
			\vspace*{-0.5em}
			\caption{
				Evaluation of state-of-the-art methods on 3DPW~\cite{3DPW}. All methods do not use 3DPW~\cite{3DPW} on training.
			}
			\vspace{-0.5 em}
			\label{tab:sota-img-video}
			
		\end{table}
		\begin{table}[!t]
			\small
			\setlength{\tabcolsep}{4.8 pt}
			\begin{center}
				\begin{tabular}{l | c | c | c | c}
					\toprule[2pt]
					\normalsize
					Module & PA-MPJPE$\downarrow$ & MPJPE$\downarrow$ & MPVPE$\downarrow$ & Accel$\downarrow$ \\
					\midrule[1pt]
					GMM & 52.6 & 84.1 & 101.0 & 6.9 \\
					\cellcolor{Gray}GMM + LPC & \cellcolor{Gray}\textbf{50.6} & \cellcolor{Gray}\textbf{80.7} & \cellcolor{Gray}\textbf{96.3} & \cellcolor{Gray}\textbf{6.6} \\
					\bottomrule[1pt]
				\end{tabular}
			\end{center}
			\vspace{-1.5 em}
			\caption{Ablation studies of Global Motion Modeling~(GMM) and Local Parameter Correction~(LPC).}
			\vspace{-1.5 em}
			\label{tab:gmmlpc}
		\end{table}
		Next, we analyze the existing problems of the previous method. 
		First, VIBE~\cite{VIBE}, TCMR~\cite{TCMR}, and MPS-Net~\cite{MPS-net} all design a single type of modeling structure to simultaneously model long-term and short-term, leading to undesirable results like global location shift and insufficient local details shown in Figure~\ref{fig:motivation}.
		Second, MEVA follows a coarse-to-fine schema to model human motion, which requires too many frames as input and is ineffective for short videos.
		
		\noindent\textbf{Single image-based and video-based methods.} We compare our GLoT with the previous single image-based and video-based methods on the challenging in-the-wild 3DPW~\cite{3DPW}. Note that all methods do not utilize the 3DPW in the training phase. 
		As shown in Table~\ref{tab:sota-img-video}, our GLoT outperforms existing single image-based and video-based methods on all metrics, which validates the effectiveness of our method.
		For example, our model surpasses the video-based MPS-Net on PA-MPJPE, MPJPE, MPVPE, and Accel by 0.5 $mm$, 1.7 $mm$, 1.8 $mm$ and 0.8 $mm/s^2$, respectively. In summary, our model achieves a smooth mesh sequence and accurate human meshes compared with previous methods. The results confirm that our proposed GLoT is effective for modeling long-range dependencies and learning local details, leading to better performance on the challenging 3DPW dataset.
		
		\subsection{Ablation studies}
		To validate the effectiveness of our GLoT, we conduct a series of experiments on 3DPW~\cite{3DPW} with the same setting as Table~\ref{tab:sota_video}.
            Initially, we verify two branches proposed in GLoT, \ie, the Global Motion Modeling and Local Parameter Correction. Subsequently, we delve into the Masked Pose and Shape Estimation strategy, in addition to analyzing the types of mask tokens employed in the Global Motion Modeling phase. Lastly, we validate the influence of varying lengths of nearby frames and the effectiveness of the Hierarchical Spatial Correlation Regressor.
		
		\begin{table}[t]
			\small
			\centering
			\setlength\tabcolsep{4pt}
			\def\arraystretch{1.0}
			
				\begin{tabular}{l|cccc}
					\toprule[2pt]
					Mask ratio~(\%) & PA-MPJPE$\downarrow$ & MPJPE$\downarrow$ & MPVPE$\downarrow$ & Accel$\downarrow$ \\
					\midrule[1pt]
					0 & 51.7 & 82.3 & 98 & 7.3  \\
					
					\cellcolor{Gray}12.5 &\cellcolor{Gray}51.3 & \cellcolor{Gray}82& \cellcolor{Gray}97.4& \cellcolor{Gray}7.1 \\
					
					25& 51.0 & 82.1 & 97.5 & 7.0  \\
					
					\cellcolor{Gray}37.5 & \cellcolor{Gray}51.0 & \cellcolor{Gray}81.4 & \cellcolor{Gray}97.2 & \cellcolor{Gray}6.8  \\
					
					50   & \textbf{50.6} & \textbf{80.7} &  \textbf{96.3} &  6.6 \\ 
					
					\cellcolor{Gray}62.5& \cellcolor{Gray}50.8 & \cellcolor{Gray}81.7 & \cellcolor{Gray}97.7 & \cellcolor{Gray}6.6 \\
					
					75  &51.4 & 82.9 & 99.1 & 6.8 \\
					
					\cellcolor{Gray}87.5  & \cellcolor{Gray}52.8 & \cellcolor{Gray}86.6 & \cellcolor{Gray}104.1 & \cellcolor{Gray}\textbf{6.5} \\
					\bottomrule[1pt]
				\end{tabular}
			\vspace*{-0.5 em}
			\caption{Ablation studies of different mask ratios.}
			\vspace{-0.5 em}
			\label{table:mask_ratio}
		\end{table}
		
		\noindent\textbf{Global Motion Modeling~(GMM) and Local Parameter Correction~(LPC).}
		We first study the GMM branch as shown in Table~\ref{tab:gmmlpc}. Our results indicate that the GMM branch alone achieves competitive results, reducing the MPJPE by 0.2~$mm$  and Accel by 0.5~$mm$ compared to MPS-Net~\cite{MPS-net}. Furthermore, our method outperforms TCMR~\cite{TCMR} in all metrics. Overall, our model shows the powerful learning long-dependency ability of the global transformer and its potential to improve intra-frame accuracy.
		Next, we combine the LPC branch into this framework, which surpasses all methods with a larger margin in terms of both intra-frame accuracy and inter-frame smoothness. It shows that our global-to-local cooperative modeling helps the model to learn coherent motion consistency and inherent human structure.

		\begin{table}[t]
			\small
			\centering
			\setlength\tabcolsep{2pt}
			\def\arraystretch{1.0}
			
				\begin{tabular}{l|c|c|c|c}
					\toprule[2pt]
					Types of mask token & PA-MPJPE$\downarrow$ & MPJPE$\downarrow$ &MPVPE$\downarrow$  & Accel$\downarrow$ \\ 
					\midrule[1pt]
					SMPL Token & \textbf{50.6} & \textbf{80.7} &  \textbf{96.3} &  \textbf{6.6} \\
					\cellcolor{Gray}Learnable Token& \cellcolor{Gray}51.5 & \cellcolor{Gray}82.5 & \cellcolor{Gray}98.5 & \cellcolor{Gray}6.9 \\ 
					\bottomrule[1pt]
				\end{tabular}
			\vspace*{-0.5 em}
			\caption{Ablation studies of types of mask tokens.}
			\vspace{-0.5 em}
			\label{tab:types of mask token}
		\end{table}
		
		\begin{table}[!t]
			\small
			\setlength{\tabcolsep}{6.5 pt}
			\begin{center}
				\begin{tabular}{l | c | c | c | c}
					\toprule[2pt]
					\normalsize
					Module & PA-MPJPE$\downarrow$ & MPJPE$\downarrow$ & MPVPE$\downarrow$ & Accel$\downarrow$ \\
					\midrule[1pt]
					Residual & 51.5 & 81.7 & 97.2 & 6.7 \\
					\cellcolor{Gray}HSCR & \cellcolor{Gray}\textbf{50.6} & \cellcolor{Gray}\textbf{80.7} & \cellcolor{Gray}\textbf{96.3} & \cellcolor{Gray}\textbf{6.6} \\
					\bottomrule[1pt]
				\end{tabular}
			\end{center}
			\vspace*{-1.5 em}
			\caption{Ablation studies of Hierarchical Spatial Correlation Regressor~(HSCR).}
			\vspace*{-1.5 em}
			\label{tab:ab_hscr}
		\end{table}

		
		\noindent\textbf{Masked Pose and Shape Estimation Strategy.}
		We randomly mask several static features during training. When testing, we do not mask any features.
		Table~\ref{table:mask_ratio} shows the influence of different mask ratios. 
		We find that (1) From the perspective of the Accel metric, applying the masked strategy helps the model to capture the long-range dependency. The Accel metric generally shows a reducing tendency when we gradually increase the mask ratio. The 87.5\% mask ratio obtains the best Accel performance, which is in line with the intuitive understanding that the model tends to learn overly smooth meshes when applying a higher mask ratio. (2) In intra-frame metrics, the mined long-term dependency also provides global contexts for the next local modeling phase. When selecting a 50\% mask ratio, our model achieves the best performance. A mask ratio between 37.5 \% to 62.5 \% is appropriate for improving the model performance.
		
		\noindent\textbf{Different types of the mask token.}
		As shown in Table~\ref{tab:types of mask token}, we draw several conclusions from the experiments with different mask tokens.
		(1) Simply applying a learnable token to masked locations obtains better performance than the previous state-of-the-art method, MPS-Net. 
		(2) When employing the SMPL token processed from the SMPL mean template, our model obtains the best result. 
		(3) We consider that applying the SMPL token complements a human mesh prior, helping the model to learn human inherent structure.
		
		\begin{table}[t]
			\small
			\setlength{\tabcolsep}{7 pt}

			\begin{center}
				\begin{tabular}{l | c c c c}
					\toprule[2pt]
					Length & PA-MPJPE$\downarrow$ & MPJPE$\downarrow$ & MPVPE$\downarrow$ & Accel$\downarrow$ \\
					\midrule[1pt]
					2 & 51.9 & 83 &	98.7 & 6.7 \\
					
					\cellcolor{Gray}3 &	\cellcolor{Gray}51.2 &	\cellcolor{Gray}84.2 & 	\cellcolor{Gray}100.0 &	\cellcolor{Gray}6.7 \\
					
					4 & \textbf{50.6} & \textbf{80.7} &  \textbf{96.3} &  \textbf{6.6} \\
					
					\cellcolor{Gray}5 & 	\cellcolor{Gray}51.3 & 	\cellcolor{Gray}81.7 &  	\cellcolor{Gray}97.2 &  	\cellcolor{Gray}6.7 \\
					
					6 & 51.7 & 82.4 &  98.4 &  6.6 \\
					\bottomrule[1pt]
				\end{tabular}
			\end{center}
			\vspace{-1.5 em}
			\caption{Ablation studies of different lengths of nearby frames. The length of 2 means the total frames feeding to the local transformer will be $2 + 2 + 1 = 5$, including the mid-frame.}
			\label{tab:nearby}
		\end{table}

		\begin{figure}[!t]
			\begin{subfigure}{0.48\linewidth}
				\includegraphics[width=1.0\textwidth]{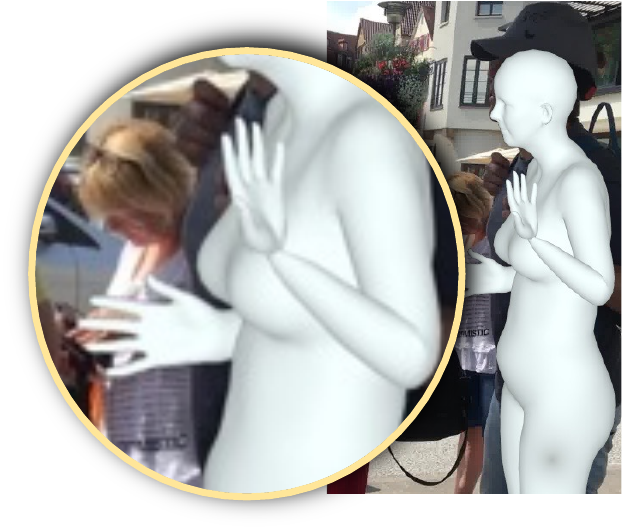}
				\vspace{-1.5 em}
				\caption{Our results of GMM. }
				\label{fig:gmm-vis}
			\end{subfigure}
			\hfill
			\begin{subfigure}{0.48\linewidth}
				\includegraphics[width=1.0\textwidth]{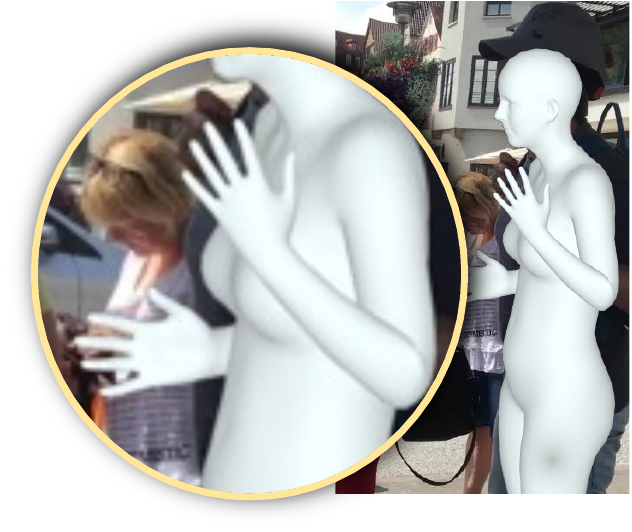}
				\vspace{-1.5 em}
				\caption{Our results via LPC. }
				\label{fig:lpc-vis}
			\end{subfigure}
			\vspace{-0.5 em}
			\caption{Qualitative results of Hierarchical Spatial Correlation Regressor (HSCR). (a) The wrist rotation is unreasonable. (b) HSCR corrects the wrist rotation. It shows that our model learns implicit kinematic constraints.}
			\label{fig:hscr_kine}
		\end{figure}
		
		\begin{figure}[!t]
			\begin{subfigure}{0.48\linewidth}
				\includegraphics[width=0.8\textwidth]{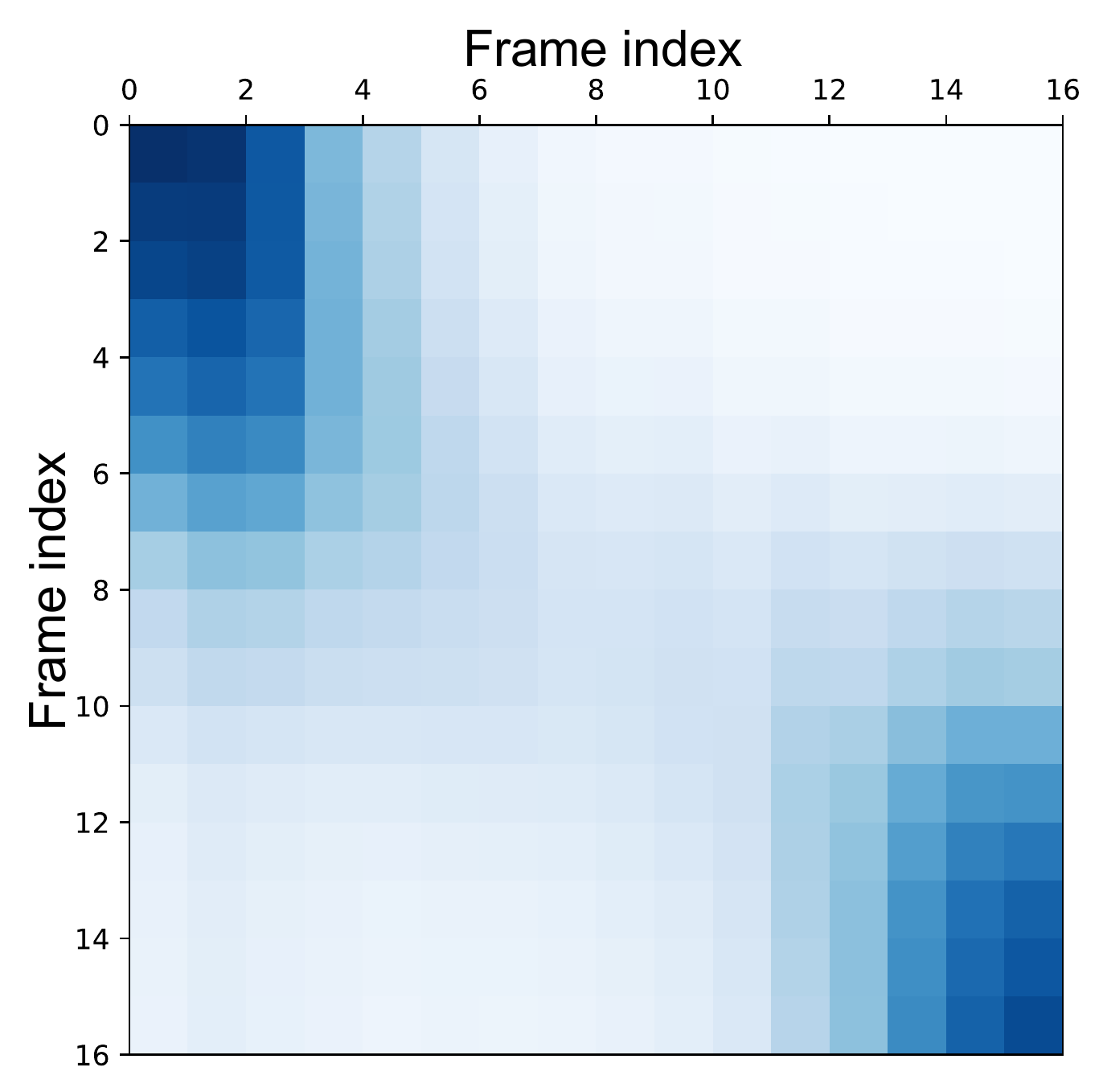}
				\centering
				\vspace{-0.5 em}
				\caption{Encoder attention weights.}
				\label{fig:en_atten_0mask}
			\end{subfigure}
			\hfill
			\begin{subfigure}{0.48\linewidth}
				\includegraphics[width=0.8\textwidth]{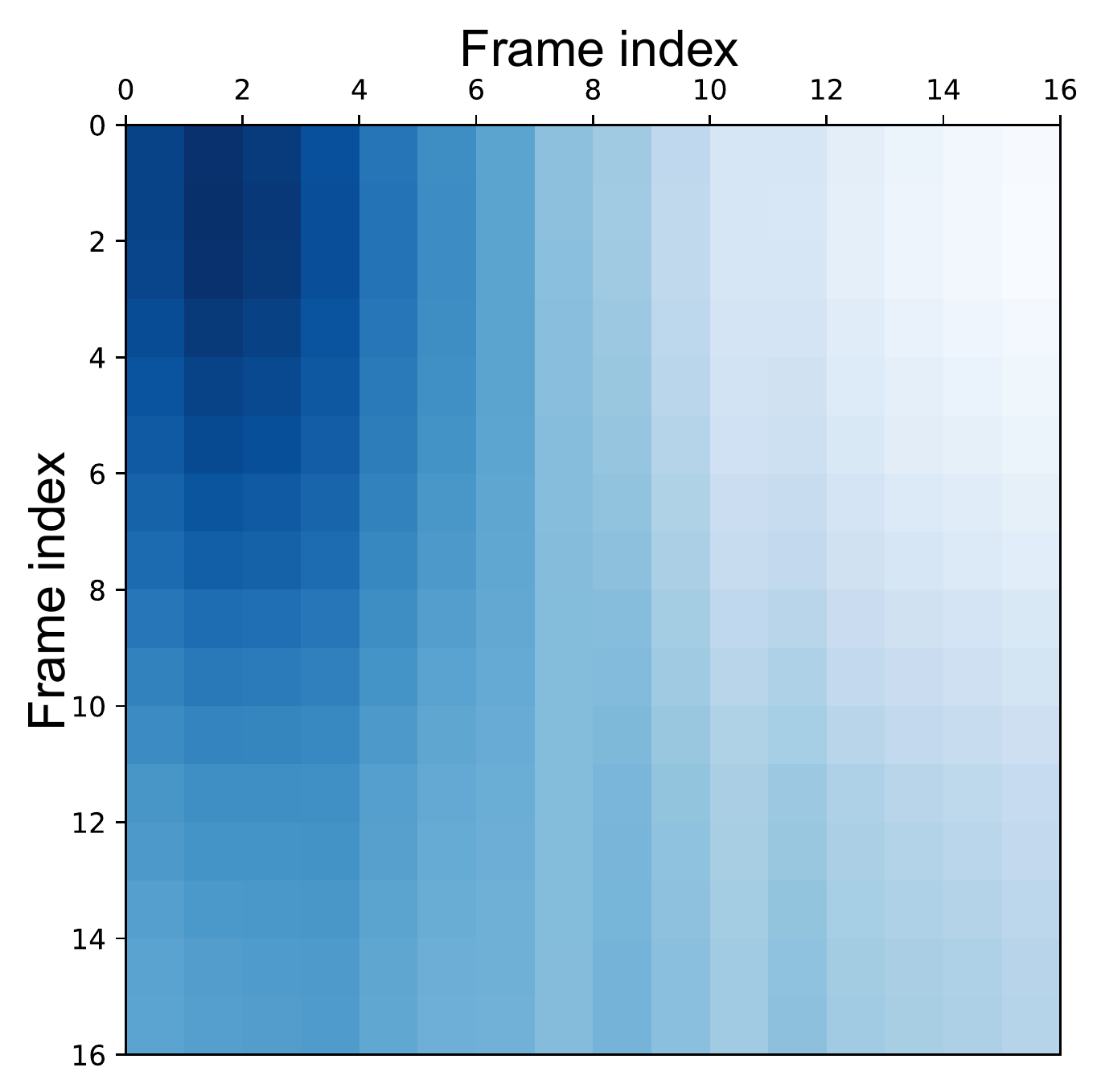}
				\centering
				\vspace{-0.5 em}
				\caption{Decoder attention weights.}
				\label{fig:de_atten_0mask}
			\end{subfigure}
			
			\begin{subfigure}{0.48\linewidth}
				\includegraphics[width=0.8\textwidth]{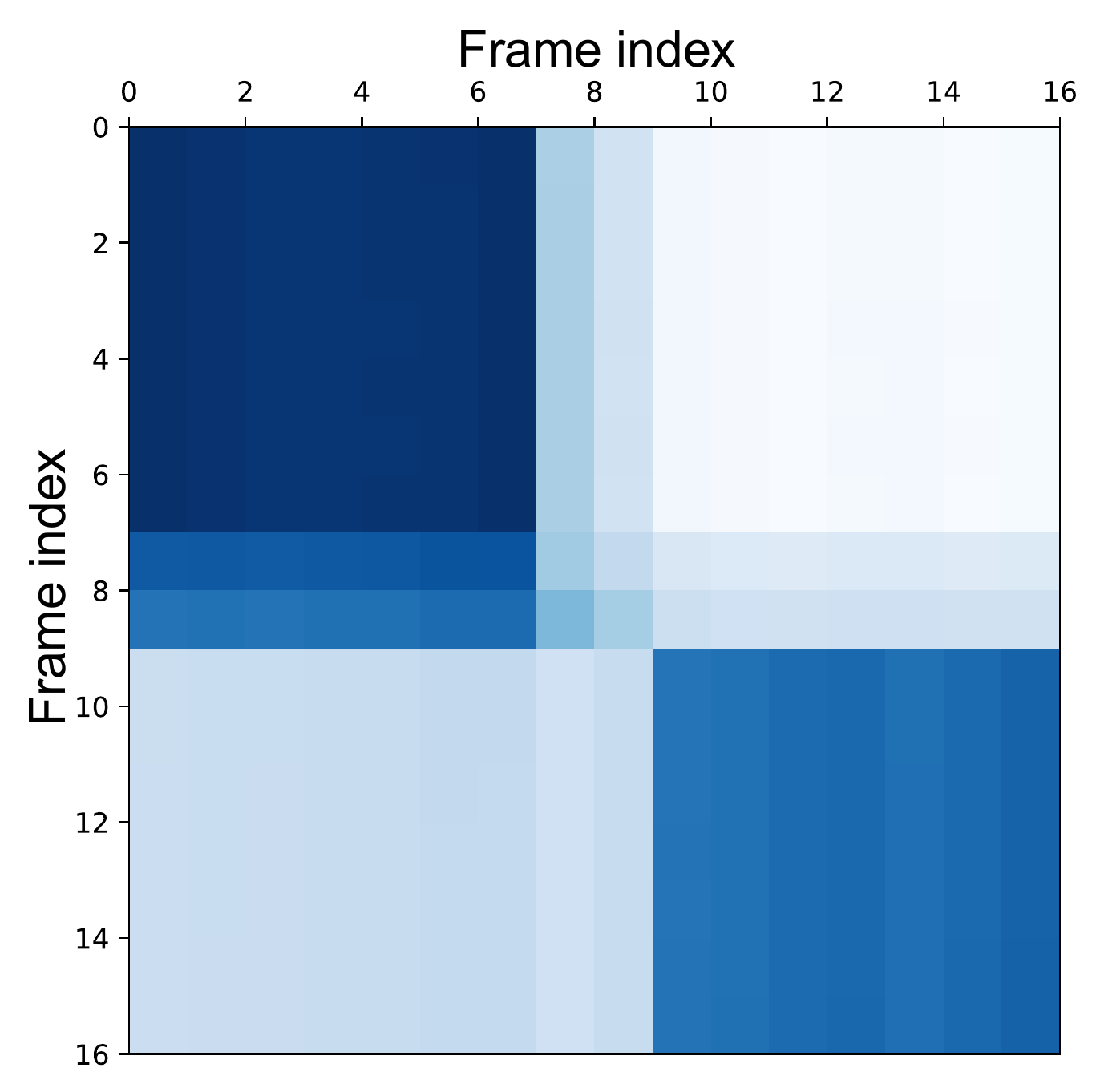}
				\centering
				\vspace{-0.5 em}
				\caption{Encoder attention weights.}
				\vspace{-0.7 em}
				\label{fig:en_atten_5mask}
			\end{subfigure}
			\hfill
			\begin{subfigure}{0.48\linewidth}
				\includegraphics[width=0.8\textwidth]{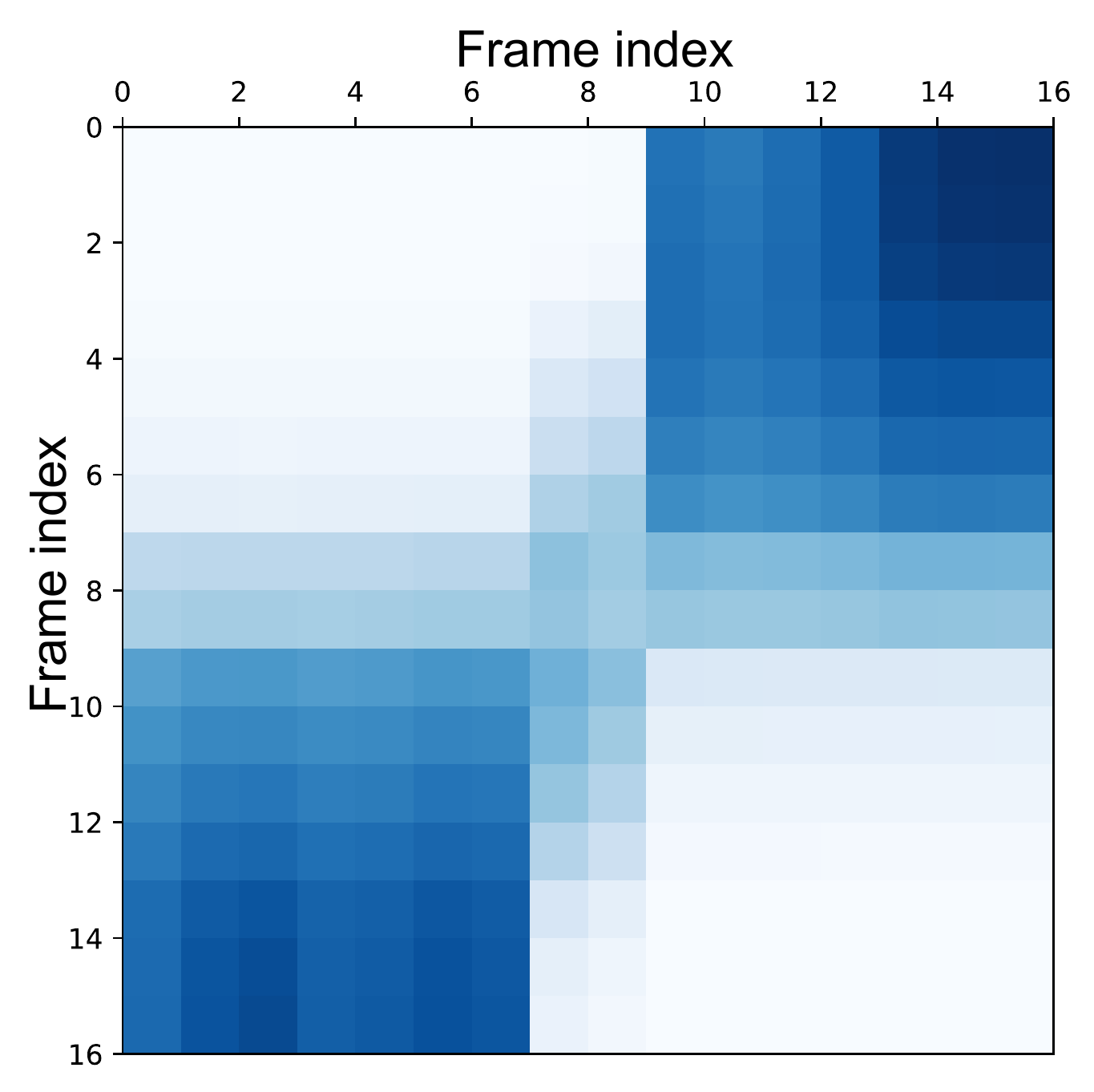}
				\centering
				\vspace{-0.5 em}
				\caption{Decoder attention weights.}
				\vspace{-0.7 em}
				\label{fig:de_atten_5mask}
			\end{subfigure}
			\caption{Attention visualizations.~(a),(b). The results without using the masking strategy. (c), (d). The results of the masking strategy.}
			\vspace{-1.5 em}
			\label{fig:atten}
		\end{figure}
		
		\begin{figure}[!t]
			\includegraphics[width=0.5\textwidth]{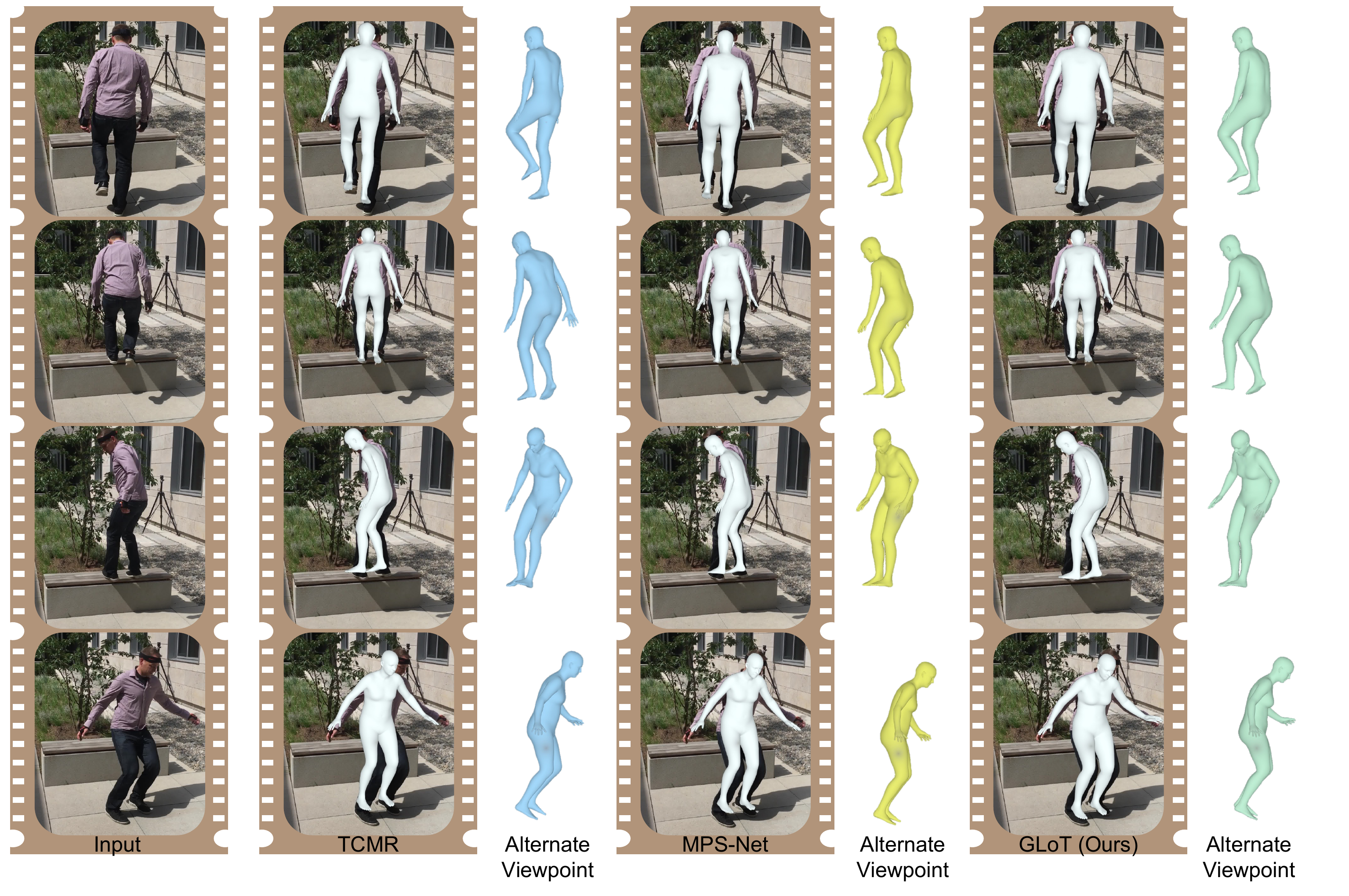}
			\vspace{-2 em}
			\caption{Qualitative comparison with previous state-of-the-art methods~\cite{TCMR, MPS-net}.}
			\vspace{-1.5 em}
			\label{fig:vis_2}
		\end{figure}

		\noindent\textbf{Hierarchical Spatial Correlation Regressor.}
		Table~\ref{tab:ab_hscr} shows the evaluation results of our HSCR. Residual means that we do not utilize the implicit kinematic constraints in Sec.~\ref{sef:hscr} and use the residual connection to replace them. We find that HSCR significantly reduces the PA-MPJPE, MPJPE, and MPVPE by 0.9 $mm$, 1.0 $mm$, and 0.9 $mm$, respectively, when compared with the residual connection. 
		Moreover, our model achieves better results compared with other previous video-based methods when merely using the residual connection.
		This indicates that our Global-to-Local framework substantially improves the performance of video-based 3D human mesh recovery.
		
		\noindent\textbf{The different lengths of nearby frames.}
		As shown in Table~\ref{tab:nearby}, we observe  the following: (1) when setting the nearby length to four, our model achieves the best performance on all metrics. (2) the length of four means the input frames of the local transformer are nine, nearly half of the input frames of the global transformer. We consider that setting it to half of the input frames of the global transformer is a good solution. (3) Although other lengths result in worse performance than four frames, they are still competitive with other methods~\cite{VIBE, TCMR, MPS-net}. For example, when the length is set to two, our method surpasses the previous state-of-the-art MPS-Net with 0.2 $mm$ PA-MPJPE, 1.3 $mm$ MPJPE, 1.0 $mm$ MPVPE, 0.7 $mm/s^2$ Accel.
		
		\subsection{Qualitative evaluation}
		
		\noindent\textbf{Hierarchical Spatial Correlation Regressor~(HSCR).}
		In Figure~\ref{fig:hscr_kine}, we study the proposed HSCR from a visualization perspective. It shows that the kinematic failure in global estimation is corrected by HSCR, which validates the effectiveness of our method.
		
		\noindent\textbf{Masked Pose and Shape Estimation strategy.}
		In Figure~\ref{fig:atten}, we provide attention visualizations of the global transformer and make several observations. (1) Compared between (a) and (c), the encoder with masking strategy attends to more nearby frames and is more centralized, indicating that it is not limited by the several nearby frames like (a). (2) The masking strategy in the decoder pushes the model to focus on further frames, while the non-masked decoder is more distributed.
	(3) The encoder and decoder in (c) and (d) show a cooperative relationship. This achieves decoupled long-term and short-term modeling.
		The encoder captures the nearby frame correlations while the decoder attends to the further frame correlations.

		\noindent\textbf{Comparison with previous methods.}
		In Figure~\ref{fig:vis_2}, we compare our method to previous approaches~\cite{TCMR, MPS-net}, and provide sequences of an alternate viewpoint. We observe that TCMR suffers global location shift and provides inaccurate meshes while MPS-Net captures the actual location, but the local details are insufficient.
		\begin{figure}[!t]
			\includegraphics[width=0.5\textwidth]{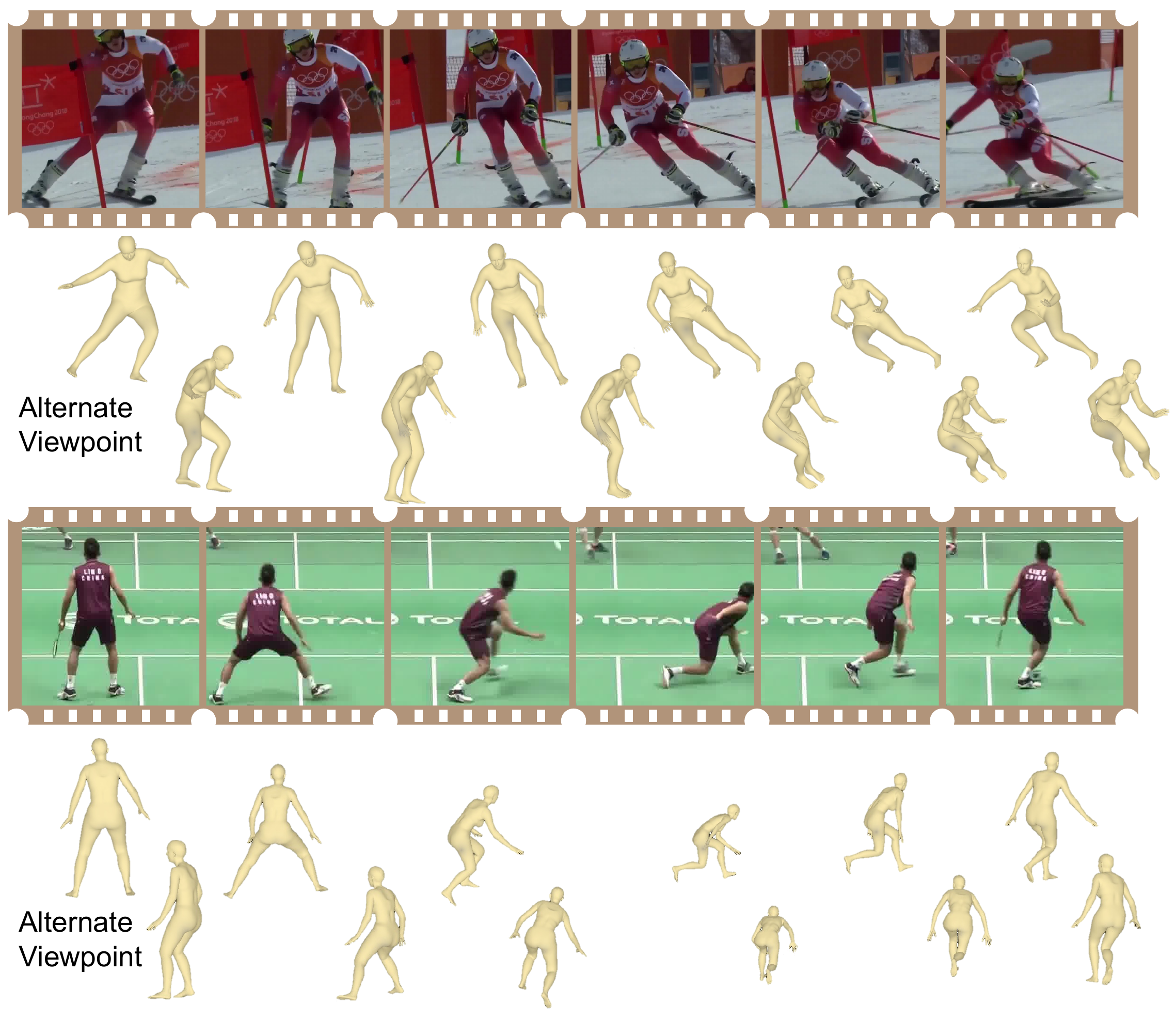}
			\vspace{-2 em}
			\caption{Qualitative results of GLoT on challenging internet videos.}
			\label{fig:vis_3}
			\vspace{-1.5 em}
		\end{figure}
		
		\noindent\textbf{Qualitative results on internet videos.}
		In Figure~\ref{fig:vis_3}, we validate our method on challenging internet videos and provide rendered meshes of an alternate viewpoint, demonstrating that our method successfully captures human motion from different perspectives. More qualitative results are provided in the supplementary material.
		
		\section{Conclusion}
		We propose a Global-to-Local Modeling~(GloT) method for video-based 3d human pose and shape estimation, which captures the long-range global dependency and local details (\eg, global location, motion consistency, and intra-frame human meshes) by combining deep networks and human prior structures.
		Through global-local cooperative modeling, GLoT achieves state-of-the-art performance on three widely used datasets.
		Furthermore, GLoT shows potential for handling in-the-wild internet videos, which could help the annotation of 3D meshes and provide various motion sequence templates for downstream tasks. 
	
\noindent\textbf{Acknowledgements.} This work was supported by the Fundamental Research Funds for the Central Universities (No. 226-2022-00051).

		{\small
			\bibliographystyle{ieee_fullname}
			\bibliography{egbib}

\begin{thebibliography}{10}\itemsep=-1pt

\bibitem{PoseTrack}
Mykhaylo Andriluka, Umar Iqbal, Eldar Insafutdinov, Leonid Pishchulin, Anton
  Milan, Juergen Gall, and Bernt Schiele.
\newblock Posetrack: A benchmark for human pose estimation and tracking.
\newblock {\em CVPR}, 2017.

\bibitem{BEIT}
Hangbo Bao, Li Dong, and Furu Wei.
\newblock Beit: Bert pre-training of image transformers.
\newblock {\em arXiv}, 2021.

\bibitem{ZheCao2018OpenPoseRM}
Zhe Cao, Gines Hidalgo, Tomas Simon, Shih-En Wei, and Yaser Sheikh.
\newblock Openpose: Realtime multi-person 2d pose estimation using part
  affinity fields.
\newblock {\em PAMI}, 2018.

\bibitem{GRU}
Kyunghyun Cho, Bart van Merri{\"e}nboer, Caglar Gulcehre, Dzmitry Bahdanau,
  Fethi Bougares, Holger Schwenk, and Yoshua Bengio.
\newblock Learning phrase representations using rnn encoder--decoder for
  statistical machine translation.
\newblock {\em empirical methods in natural language processing}, 2014.

\bibitem{TCMR}
Hongsuk Choi, Gyeongsik Moon, Ju~Yong Chang, and Kyoung~Mu Lee.
\newblock Beyond static features for temporally consistent 3d human pose and
  shape from a video.
\newblock {\em CVPR}, 2020.

\bibitem{BERT}
Jacob Devlin, Ming-Wei Chang, Kenton Lee, and Kristina Toutanova.
\newblock Bert: Pre-training of deep bidirectional transformers for language
  understanding.
\newblock {\em north american chapter of the association for computational
  linguistics}, 2018.

\bibitem{VIT}
Alexey Dosovitskiy, Lucas Beyer, Alexander Kolesnikov, Dirk Weissenborn,
  Xiaohua Zhai, Thomas Unterthiner, Mostafa Dehghani, Matthias Minderer, Georg
  Heigold, Sylvain Gelly, Jakob Uszkoreit, and Neil Houlsby.
\newblock An image is worth 16x16 words: Transformers for image recognition at
  scale.
\newblock {\em arXiv}, 2020.

\bibitem{MatthijsDouze2020TrainingDI}
Matthijs Douze, Hugo Touvron, Matthieu Cord, Douze Matthijs, Francisco Massa,
  Alexandre Sablayrolles, and Herv{\'e} J{\'e}gou.
\newblock Training data-efficient image transformers \& distillation through
  attention.
\newblock {\em international conference on machine learning}, 2020.

\bibitem{GeorgiosGeorgakis2020HierarchicalKH}
Georgios Georgakis, Ren Li, Srikrishna Karanam, Terrence Chen, Jana Kosecka,
  and Ziyan Wu.
\newblock Hierarchical kinematic human mesh recovery.
\newblock {\em ECCV}, 2020.

\bibitem{MAE}
Kaiming He, Xinlei Chen, Saining Xie, Yanghao Li, Piotr Doll{\'a}r, and Ross
  Girshick.
\newblock Masked autoencoders are scalable vision learners.
\newblock {\em CVPR}, 2021.

\bibitem{ResNet}
Kaiming He, Xiangyu Zhang, Shaoqing Ren, and Jian Sun.
\newblock Deep residual learning for image recognition.
\newblock {\em CVPR}, 2015.

\bibitem{H36M}
Catalin Ionescu, Dragos Papava, Vlad Olaru, and Cristian Sminchisescu.
\newblock Human3.6m: Large scale datasets and predictive methods for 3d human
  sensing in natural environments.
\newblock {\em PAMI}, 2014.

\bibitem{HMR}
Angjoo Kanazawa, Michael~J. Black, David~W. Jacobs, and Jitendra Malik.
\newblock End-to-end recovery of human shape and pose.
\newblock In {\em CVPR}, 2018.

\bibitem{Insta}
Angjoo Kanazawa, Jason~Y. Zhang, Panna Felsen, and Jitendra Malik.
\newblock Learning 3d human dynamics from video.
\newblock {\em CVPR}, 2018.

\bibitem{DiederikPKingma2014AdamAM}
Diederik~P. Kingma and Jimmy Ba.
\newblock Adam: A method for stochastic optimization.
\newblock {\em arXiv}, 2014.

\bibitem{VAE}
Diederik~P. Kingma and Max Welling.
\newblock Auto-encoding variational bayes.
\newblock {\em arXiv}, 2013.

\bibitem{VIBE}
Muhammed Kocabas, Nikos Athanasiou, and Michael~J. Black.
\newblock Vibe: Video inference for human body pose and shape estimation.
\newblock {\em CVPR}, 2020.

\bibitem{PARE}
Muhammed Kocabas, Chun-Hao~P Huang, Otmar Hilliges, and Michael~J Black.
\newblock Pare: Part attention regressor for 3d human body estimation.
\newblock In {\em ICCV}, pages 11127--11137, 2021.

\bibitem{SPIN}
Nikos Kolotouros, Georgios Pavlakos, Michael~J Black, and Kostas Daniilidis.
\newblock Learning to reconstruct 3d human pose and shape via model-fitting in
  the loop.
\newblock In {\em ICCV}, 2019.

\bibitem{GraphCMR}
Nikos Kolotouros, Georgios Pavlakos, and Kostas Daniilidis.
\newblock Convolutional mesh regression for single-image human shape
  reconstruction.
\newblock In {\em CVPR}, pages 4501--4510, 2019.

\bibitem{li2021hybrik}
Jiefeng Li, Chao Xu, Zhicun Chen, Siyuan Bian, Lixin Yang, and Cewu Lu.
\newblock Hybrik: A hybrid analytical-neural inverse kinematics solution for 3d
  human pose and shape estimation.
\newblock In {\em CVPR}, pages 3383--3393, 2021.

\bibitem{li2022locality}
Liulei Li, Tianfei Zhou, Wenguan Wang, Lu Yang, Jianwu Li, and Yi Yang.
\newblock Locality-aware inter-and intra-video reconstruction for
  self-supervised correspondence learning.
\newblock In {\em CVPR}, pages 8719--8730, 2022.

\bibitem{liang2023local}
Chen Liang, Wenguan Wang, Tianfei Zhou, Jiaxu Miao, Yawei Luo, and Yi Yang.
\newblock Local-global context aware transformer for language-guided video
  segmentation.
\newblock {\em PAMI}, 2023.

\bibitem{metro_lin}
Kevin Lin, Lijuan Wang, and Zicheng Liu.
\newblock End-to-end human pose and mesh reconstruction with transformers.
\newblock In {\em CVPR}, pages 1954--1963, 2021.

\bibitem{ZeLiu2021SwinTH}
Ze Liu, Yutong Lin, Yue Cao, Han Hu, Yixuan Wei, Zheng Zhang, Stephen Lin, and
  Baining Guo.
\newblock Swin transformer: Hierarchical vision transformer using shifted
  windows.
\newblock {\em ICCV}, 2021.

\bibitem{Mosh}
Matthew Loper, Naureen Mahmood, and Michael~J Black.
\newblock Mosh: motion and shape capture from sparse markers.
\newblock {\em ACM Trans. Graph.}, 33(6):220--1, 2014.

\bibitem{SMPL}
Matthew Loper, Naureen Mahmood, Javier Romero, Gerard Pons-Moll, and Michael~J
  Black.
\newblock Smpl: A skinned multi-person linear model.
\newblock {\em TOG}, 34(6):1--16, 2015.

\bibitem{IlyaLoshchilov2016SGDRSG}
Ilya Loshchilov and Frank Hutter.
\newblock Sgdr: Stochastic gradient descent with warm restarts.
\newblock {\em arXiv}, 2016.

\bibitem{MEVA}
Zhengyi Luo, S.~Alireza Golestaneh, and Kris~M. Kitani.
\newblock 3d human motion estimation via motion compression and refinement.
\newblock {\em ACCV}, 2020.

\bibitem{AMASS}
Naureen Mahmood, Nima Ghorbani, Nikolaus~F. Troje, Gerard Pons-Moll, and
  Michael~J. Black.
\newblock Amass: Archive of motion capture as surface shapes.
\newblock {\em ICCV}, 2019.

\bibitem{MPII3D}
Dushyant Mehta, Helge Rhodin, Dan Casas, Pascal Fua, Oleksandr Sotnychenko,
  Weipeng Xu, and Christian Theobalt.
\newblock Monocular 3d human pose estimation in the wild using improved cnn
  supervision.
\newblock {\em international conference on 3d vision}, 2016.

\bibitem{miao2022large}
Jiaxu Miao, Xiaohan Wang, Yu Wu, Wei Li, Xu Zhang, Yunchao Wei, and Yi Yang.
\newblock Large-scale video panoptic segmentation in the wild: A benchmark.
\newblock In {\em CVPR}, pages 21033--21043, 2022.

\bibitem{miao2021vspw}
Jiaxu Miao, Yunchao Wei, Yu Wu, Chen Liang, Guangrui Li, and Yi Yang.
\newblock Vspw: A large-scale dataset for video scene parsing in the wild.
\newblock In {\em CVPR}, pages 4133--4143, 2021.

\bibitem{moon2020i2l}
Gyeongsik Moon and Kyoung~Mu Lee.
\newblock I2l-meshnet: Image-to-lixel prediction network for accurate 3d human
  pose and mesh estimation from a single rgb image.
\newblock In {\em ECCV}, pages 752--768. Springer, 2020.

\bibitem{MohamedOmran2018NeuralBF}
Mohamed Omran, Christoph Lassner, Gerard Pons-Moll, Peter~V. Gehler, and Bernt
  Schiele.
\newblock Neural body fitting: Unifying deep learning and model based human
  pose and shape estimation.
\newblock {\em international conference on 3d vision}, 2018.

\bibitem{GeorgiosPavlakos2016CoarsetoFineVP}
Georgios Pavlakos, Xiaowei Zhou, Konstantinos~G. Derpanis, and Kostas
  Daniilidis.
\newblock Coarse-to-fine volumetric prediction for single-image 3d human pose.
\newblock {\em CVPR}, 2016.

\bibitem{GeorgiosPavlakos2018LearningTE}
Georgios Pavlakos, Luyang Zhu, Xiaowei Zhou, and Kostas Daniilidis.
\newblock Learning to estimate 3d human pose and shape from a single color
  image.
\newblock {\em CVPR}, 2018.

\bibitem{JavierRomero2017EmbodiedHM}
Javier Romero, Dimitrios Tzionas, and Michael~J. Black.
\newblock Embodied hands: modeling and capturing hands and bodies together.
\newblock {\em TOG}, 2017.

\bibitem{XiaolongShen2022StepNetSP}
Xiaolong Shen, Zhedong Zheng, and Yi Yang.
\newblock Stepnet: Spatial-temporal part-aware network for sign language
  recognition.
\newblock In {\em arXiv}, 2022.

\bibitem{Transformer}
Ashish Vaswani, Noam Shazeer, Niki Parmar, Jakob Uszkoreit, Llion Jones,
  Aidan~N. Gomez, Lukasz Kaiser, and Illia Polosukhin.
\newblock Attention is all you need.
\newblock {\em NIPS}, 2017.

\bibitem{3DPW}
Timo von Marcard, Roberto Henschel, Michael~J. Black, Bodo Rosenhahn, and
  Gerard Pons-Moll.
\newblock Recovering accurate \{3D\} human pose in the wild using \{IMUs\} and
  a moving camera.
\newblock {\em ECCV}, 2018.

\bibitem{ZiniuWan2021EncoderDecoderWM}
Ziniu Wan, Zhengjia Li, Tian Maoqing, Jianbo Liu, Shuai Yi, and Hongsheng Li.
\newblock Encoder-decoder with multi-level attention for 3d human shape and
  pose estimation.
\newblock {\em ICCV}, 2021.

\bibitem{Non_local}
Xiaolong Wang, Ross Girshick, Abhinav Gupta, and Kaiming He.
\newblock Non-local neural networks.
\newblock {\em CVPR}, 2017.

\bibitem{MPS-net}
Wen-Li Wei, Jen-Chun Lin, Tyng-Luh Liu, and Hong-Yuan~Mark Liao.
\newblock Capturing humans in motion: Temporal-attentive 3d human pose and
  shape estimation from monocular video.
\newblock In {\em CVPR}, pages 13211--13220, 2022.

\bibitem{yang2021multiple}
Yi Yang, Yueting Zhuang, and Yunhe Pan.
\newblock Multiple knowledge representation for big data artificial
  intelligence: framework, applications, and case studies.
\newblock {\em Frontiers of Information Technology \& Electronic Engineering},
  22(12):1551--1558, 2021.

\bibitem{yang2021associating}
Zongxin Yang, Yunchao Wei, and Yi Yang.
\newblock Associating objects with transformers for video object segmentation.
\newblock In {\em NIPS}, volume~34, pages 2491--2502, 2021.

\bibitem{yang2021collaborative}
Zongxin Yang, Yunchao Wei, and Yi Yang.
\newblock Collaborative video object segmentation by multi-scale
  foreground-background integration.
\newblock {\em PAMI}, 44(9):4701--4712, 2021.

\bibitem{yang2022decoupling}
Zongxin Yang and Yi Yang.
\newblock Decoupling features in hierarchical propagation for video object
  segmentation.
\newblock In {\em NIPS}, 2022.

\bibitem{T2T}
Li Yuan, Yunpeng Chen, Tao Wang, Weihao Yu, Yujun Shi, Zi-Hang Jiang,
  Francis~EH Tay, Jiashi Feng, and Shuicheng Yan.
\newblock Tokens-to-token vit: Training vision transformers from scratch on
  imagenet.
\newblock In {\em ICCV}, pages 558--567, 2021.

\bibitem{zhang2021pymaf}
Hongwen Zhang, Yating Tian, Xinchi Zhou, Wanli Ouyang, Yebin Liu, Limin Wang,
  and Zhenan Sun.
\newblock Pymaf: 3d human pose and shape regression with pyramidal mesh
  alignment feedback loop.
\newblock In {\em ICCV}, pages 11446--11456, 2021.

\bibitem{Penn}
Weiyu Zhang, Menglong Zhu, and Konstantinos~G. Derpanis.
\newblock From actemes to action: A strongly-supervised representation for
  detailed action understanding.
\newblock {\em ICCV}, 2013.

\bibitem{zhao2022centerclip}
Shuai Zhao, Linchao Zhu, Xiaohan Wang, and Yi Yang.
\newblock Centerclip: Token clustering for efficient text-video retrieval.
\newblock In {\em Proceedings of the 45th International ACM SIGIR Conference on
  Research and Development in Information Retrieval}, pages 970--981, 2022.

\bibitem{zhou2022survey}
Tianfei Zhou, Fatih Porikli, David~J Crandall, Luc Van~Gool, and Wenguan Wang.
\newblock A survey on deep learning technique for video segmentation.
\newblock {\em PAMI}, 2022.

\bibitem{zimmermann2019freihand}
Christian Zimmermann, Duygu Ceylan, Jimei Yang, Bryan Russell, Max Argus, and
  Thomas Brox.
\newblock Freihand: A dataset for markerless capture of hand pose and shape
  from single rgb images.
\newblock In {\em CVPR}, pages 813--822, 2019.

\end{thebibliography}
		}
\section{
		Appendix
	}
\appendix
\section{Datasets}
	\begin{figure}[!t]
		\begin{subfigure}{0.48\linewidth}
			\includegraphics[width=1.0\textwidth]{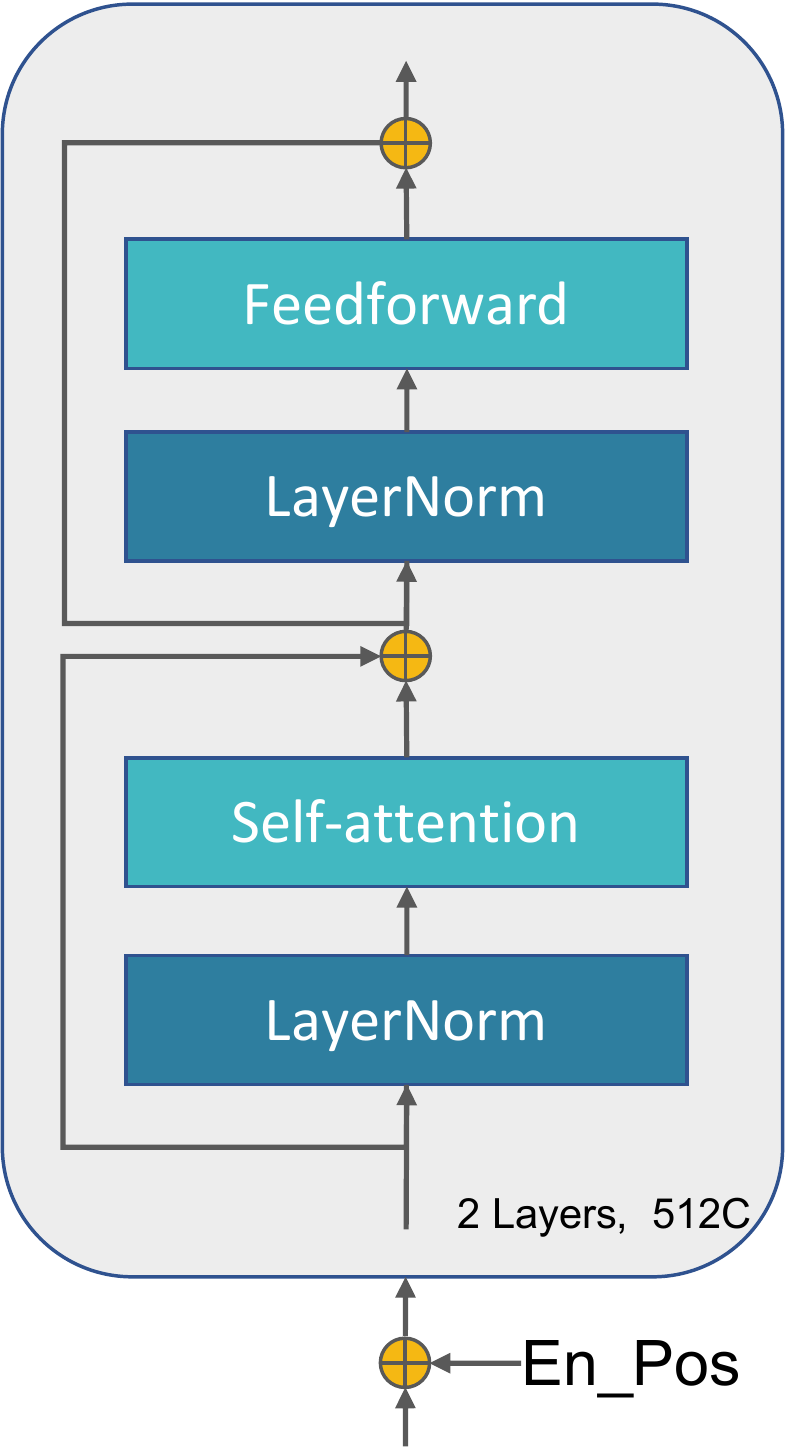}
			\caption{Global encoder.}
		\end{subfigure}
		\hfill
		\begin{subfigure}{0.48\linewidth}
			\includegraphics[width=1.0\textwidth]{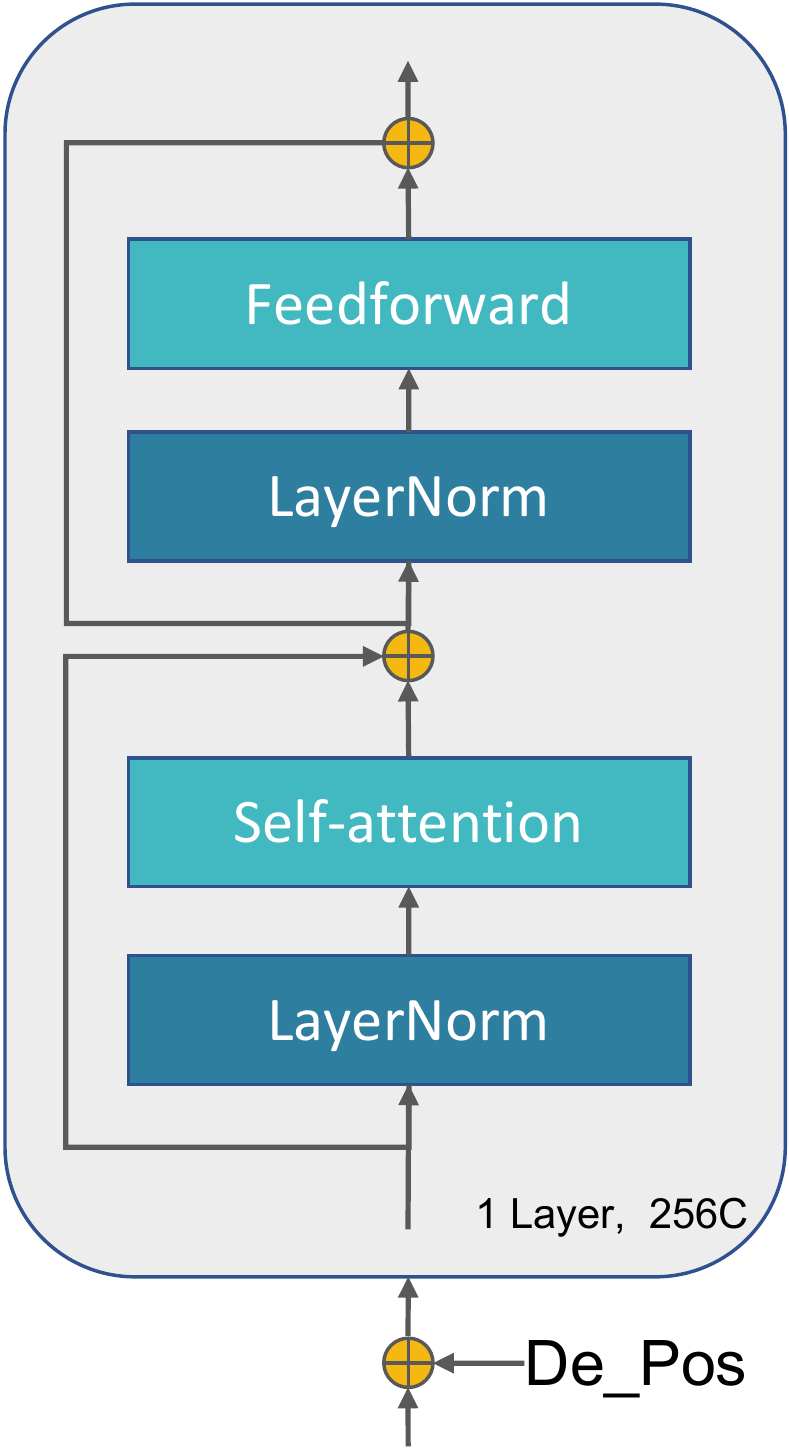}
			\caption{Global decoder.}
		\end{subfigure}
		\caption{Details of the global transformer.}
		\vspace{-1.5 em}
		\label{fig:global_transformer}
	\end{figure}
	
	\noindent\textbf{3DPW.}
	3DPW~\cite{3DPW} is a challenging in-the-wild consisting of 60 videos, which are captured by a phone at 30 fps. Moreover, IMU sensors are utilized to obtain the near ground-truth SMPL parameters, \ie, pose and shape. We utilize the official split to train and test our model, where the training, validation, and test sets are comprised of 24, 12, and 24 videos, respectively. For evaluation, we report MPVPE on 3DPW because it has ground-truth shape annotation.

	\noindent\textbf{Human3.6M.}
	Human3.6M~\cite{H36M} is a large-scale dataset collected under a controlled indoor environment and includes 3.6M video frames. Folloing~\cite{TCMR,MPS-net}, we train the model on 5 subjects (i.e., S1, S5, S6, S7, and S8) and test it on 2 subjects (i.e., S9 and S11). We set the frame rate of the dataset to 25 fps for training and testing.
	
	\noindent\textbf{MPI-INF-3DHP.}
    MPI-INF-3DHP~\cite{MPII3D}] is a complex dataset captured at indoor and outdoor scenes with a markerless motion capture system. The 3D human pose annotations are computed by the multiview method.
    The training and testing sets are comprised of 8 and 6 subjects, respectively. Each subject has 16 videos captured in the indoor or outdoor environment. The total video frames are 1.3M.
    Following previous works~\cite{TCMR, MPS-net}, we utilize the official training and testing split. 

	\noindent\textbf{InstaVariety.}
	InstaVariety is a 2D human pose dataset collected from Instagram. It consists of 28K videos, and the video length is an average of 6 seconds. The 2D annotation is generated from Openpose~\cite{ZheCao2018OpenPoseRM}. Following~\cite{TCMR, MPII3D}, we use this dataset for training.

	\noindent\textbf{PoseTrack.}
    PoseTrack~\cite{PoseTrack} is also a 2D human dataset for multi-person pose estimation and tracking, which consists of 1.3K videos. Following~\cite{TCMR}, we use 792 videos for training. 
    
    \begin{figure}[!t]
		\includegraphics[width=0.4\textwidth]{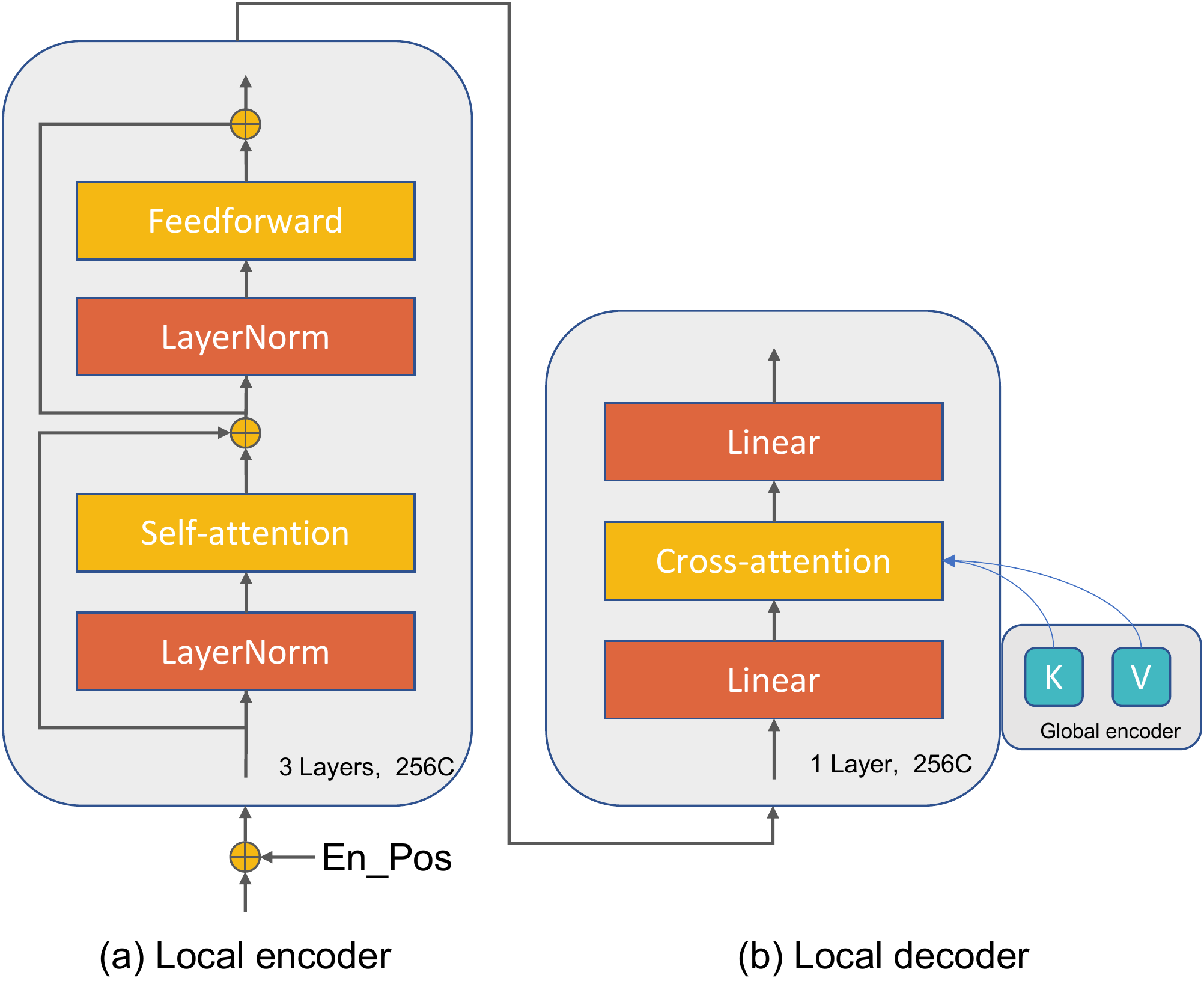}
		\caption{Details of the local transformer.}
		\label{fig:local_transformer}
	\end{figure}
	
	\begin{table}[!t]
		\small
		\setlength{\tabcolsep}{5. pt}
		\begin{center}
			\begin{tabular}{l | c | c | c | c}
				\toprule[2pt]
				\normalsize
				 & PA-MPJPE$\downarrow$ & MPJPE$\downarrow$ & MPVPE$\downarrow$ & Accel$\downarrow$ \\
				\midrule[1pt]
				 $w/o~Detach$ & 50.9 & 81.2 & 96.4 & 6.6 \\
				\cellcolor{Gray}$w/~Detach$ & \cellcolor{Gray}\textbf{50.6} & \cellcolor{Gray}\textbf{80.7} & \cellcolor{Gray}\textbf{96.3} & \cellcolor{Gray}\textbf{6.6} \\
				\bottomrule[1pt]
			\end{tabular}
		\end{center}
		\vspace*{-1.8 em}
		\caption{Gradient detachment}
		\vspace*{-.5 em}
		\label{tab:gradient}
	\end{table}

	\begin{table}[!t]
		\vspace*{-.5 em}
		\small
		\setlength{\tabcolsep}{6. pt}
		\begin{center}
			\begin{tabular}{l | c | c | c | c}
				\toprule[2pt]
				\normalsize
				 & PA-MPJPE$\downarrow$ & MPJPE$\downarrow$ & MPVPE$\downarrow$ & Accel$\downarrow$ \\
				\midrule[1pt]
				Fix & 52.1 & 83.4 & 99.5 & 6.4 \\
				\cellcolor{Gray}Learnable& \cellcolor{Gray}\textbf{50.6} & \cellcolor{Gray}\textbf{80.7} & \cellcolor{Gray}\textbf{96.3} & \cellcolor{Gray}\textbf{6.6} \\
				\bottomrule[1pt]
			\end{tabular}
		\end{center}
		\vspace*{-1.8 em}
		\caption{Position embedding}
		\vspace*{-2.2em}
		\label{tab:position}
	\end{table}

\begin{figure*}[!t]
    \centering
    \begin{subfigure}{0.32\linewidth}
    \animategraphics[width=5.4cm,height=5.68cm, autoplay, loop]{2}{gif_1/our/000}{01}{16}
        \caption{\small Ours}
        
    \end{subfigure}
    \begin{subfigure}{0.32\linewidth}
    \animategraphics[width=5.4cm,height=5.68cm, autoplay, loop]{2}{gif_1/mps/000}{01}{16}
        \caption{\small MPS-Net~\cite{MPS-net}}
        
    \end{subfigure}
    \begin{subfigure}{0.32\linewidth}
    \animategraphics[width=5.4cm,height=5.68cm, autoplay, loop]{2}{gif_1/tcmr/000}{01}{16}
        \caption{\small TCMR~\cite{TCMR}}
        
    \end{subfigure}
    \vspace{-0.5 em}
    \caption{Comparison with other methods~\cite{TCMR, MPS-net}. \textbf{Please use Adobe Acrobat to view it.}}
    \label{fig:more_1}
    \vspace{-0.5 em}
\end{figure*}

\begin{figure*}[!t]
    \centering
    \animategraphics[width=17cm,height=4.8cm, autoplay, loop]{2}{gif_2/0000}{00}{20}
    \vspace{-0.5 em}
    \caption{An Example of internet video. We sample every ten frames. \textbf{Please use Adobe Acrobat to view it.}}
    \label{fig:more_2}
    \vspace{-0.5 em}
\end{figure*}

\section{Model details}
    \noindent\textbf{Global transformer.}
    The model details are shown in Figure~\ref{fig:global_transformer}. We utilize two layers of the encoder block with 512 dimensions. In the global decoder, we only apply one layer of the decoder block with 256 dimensions. The position embedding is learnable.

	\noindent\textbf{Local transformer.}
	As shown in Figure~\ref{fig:local_transformer}, the encoder block is similar to the global encoder. We set three layers of the encoder block with 256 channel sizes. In addition, we employ cross-attention to the decoder and set the layer to one. The channel size is the same as the encoder.

\begin{figure*}[!t]
    \centering
    \animategraphics[width=17cm,height=4.8cm, autoplay, loop]{2}{gif_3/00}{01}{19}
    \vspace{-0.5 em}
    \caption{An Example of internet video. \textbf{Please use Adobe Acrobat to view it.}}
    \vspace{-0.5 em}
    \label{fig:more_3}
\end{figure*}
\section{Effect of gradient detachment}
Table~\ref{tab:gradient} shows the effect of gradient detachment. 
When we do not backward propagate the path of global estimation to HSCR, GLoT achieves the best performance. It is intuitively reasonable that fixing one is easier for optimization. In addition, $w/o~Detach$ also obtains good results. 

\section{Effect of position embedding}
In Table~\ref{tab:position}, we report the results of the different types of position embeddings. The learnable embedding obtains the best performance. 

\begin{figure}[!t]
		\begin{subfigure}{0.625\linewidth}
			\includegraphics[width=1.0\textwidth]{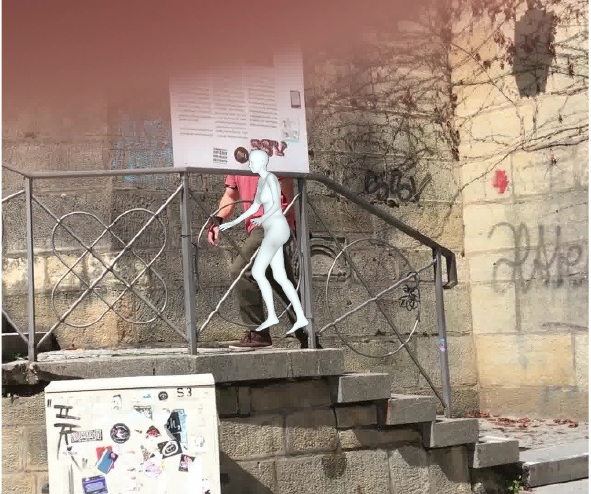}
			\vspace{-1.5 em}
			\label{fig:failure_1}
		\end{subfigure}
		\hfill
		\begin{subfigure}{0.335\linewidth}
			\includegraphics[width=1.0\textwidth]{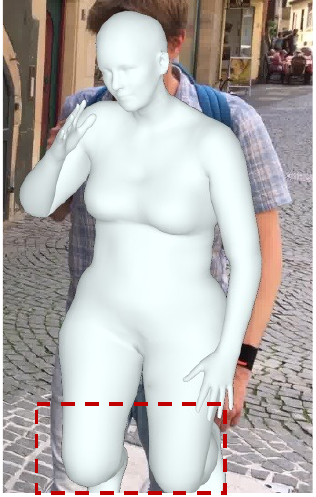}
			\vspace{-1.5 em}
			\label{fig:failure_2}
		\end{subfigure}
		\vspace{-0.5 em}
		\caption{Some failure cases.}
		\vspace{-0.5 em}
		\label{fig:failure}
\end{figure}

\section{Inference time (GPU: V100) and MACs}

\begin{table}[htb]
	\small
	\vspace*{-0.75 em}
	\setlength{\tabcolsep}{6 pt}
	\begin{center}
		\begin{tabular}{l | c | c | c }
			\toprule[2pt]
			\normalsize
			Model & MACs (M) & Time (ms) & PA-MPJPE\\
			\midrule[1pt]
			TCMR & 861.8 & \textbf{11.7} & 52.7\\
			\cellcolor{Gray}MPS-Net & \cellcolor{Gray}318.4 & \cellcolor{Gray}17.6  & \cellcolor{Gray}52.1\\
			Ours w/ Residual & \textbf{287.9} & 13.0 & 51.5\\
			\cellcolor{Gray}Ours w/ HSCR & \cellcolor{Gray}288.1 & \cellcolor{Gray}16.2 & \cellcolor{Gray}\textbf{50.6}\\
			\bottomrule[1pt]
		\end{tabular}
	\end{center}
	\vspace*{-1.5 em}
	\caption{Inference time and MACs.}
	\vspace*{-1. em}
	\label{tab:infer}
\end{table}

We provide the results of inference time and MACs in Table~\ref{tab:infer}. 
Our model achieves the lowest MACs. 
For inference time, our model (w/ HSCR) is slower than TCMR~\cite{TCMR} but faster than the previous SOTA method MPS-Net~\cite{MPS-net}. We analyze that the reason for slower than TCMR is the self-attention mechanism used in our model and MPS-Net.
Although our model (w/ HSCR) is slower than TCMR by 4.5 ms, it shows a significant improvement in PA-MPJPE.
Besides, we provide the inference time of our model (w/ Residual) for comparing the time consumption of the HSCR. 
It is worth noting that our model (w/ Residual) reduces 1.2 PA-MPJPE with a time consumption of only 1.3 ms compared with TCMR.

\section{Input length of the global encoder}
\begin{table}[htb]
	\small
	\vspace*{-1.5 em}
	\setlength{\tabcolsep}{6.5 pt}
	\begin{center}
		\begin{tabular}{l | c | c | c | c}
			\toprule[2pt]
			\normalsize
			length & PA-MPJPE$\downarrow$ & MPJPE$\downarrow$ & MPVPE$\downarrow$ & Accel$\downarrow$ \\
			\midrule[1pt]
			32 & 51.2 & 82.0 & 98.3 & 6.7 \\
			\cellcolor{Gray}24 &
			\cellcolor{Gray}51.2 & \cellcolor{Gray}82.7 & \cellcolor{Gray}98.5 & \cellcolor{Gray}6.7 \\
			16 & \textbf{50.6} & \textbf{80.7} &
			\textbf{96.3} & 
			\textbf{6.6} \\
			\bottomrule[1pt]
		\end{tabular}
	\end{center}
	\vspace*{-1.5 em}
	\caption{Input length of the global encoder.}
	\vspace*{-1. em}
	\label{input_length}
	
\end{table}

Although the 16-frame input length is commonly used in this task, we consider that exploring more input lengths is valuable.
In Table~\ref{input_length}, we supply the study of longer input lengths, 24 and 32.
The 16-frame setting achieves the best results. A possible reason is that our lightweight global encoder can not sufficiently model longer temporal relations. 

\section{More qualitative results}
We show the comparison results with other methods in Figure~\ref{fig:more_1}. We observe (1) The results of MPS-Net~\cite{MPS-net} suffer from insufficient local details. (2) The results of TCMR~\cite{TCMR} do not capture the actual human global location of the frames. Figure~\ref{fig:more_2} and~\ref{fig:more_3} are multi-person internet videos, we first use a multi-object tracker to process videos and then utilize our method for each tracked person, following the previous methods~\cite{TCMR, MPS-net}.

\section{Failure cases}

As shown in Figure~\ref{fig:failure}, we provide some failure cases, mainly including occlusion. We divide the occlusion into two types, \ie, object occlusion~(Left Figure) and truncation of the frame~(Right Figure, some joints are outside of the frame). 
We consider that these cases are caused by long-term occlusion, which means the input frames are all occluded by the object or truncated by the camera, leading to failures in temporal modeling. 

\section{Future works}
We plan to use this framework in similar tasks, \ie, hand pose and shape estimation~\cite{JavierRomero2017EmbodiedHM, zimmermann2019freihand}. This task will provide a more robust hand representation for downstream tasks, \eg, sign language recognition~\cite{XiaolongShen2022StepNetSP}.
Moreover, we believe that exploring multi-person interaction in a video would be a good idea. While there are some methods in image-based tasks to deal with occlusion problems caused by multiple people, video-based methods in this area are still unexplored.



	\end{document}